\documentclass[preprint,12pt]{elsarticle}

\usepackage{amssymb}

\usepackage{graphicx}
\usepackage{latexsym}
\usepackage{caption}
\usepackage{url}
\usepackage{subcaption}
\usepackage{amssymb}
\usepackage{lipsum}
\usepackage{graphicx}
\usepackage{array}
\usepackage{longtable}
\usepackage{soul}
\usepackage{comment}
\usepackage{amsmath}
\usepackage{algorithm}
\usepackage{algpseudocode}
\usepackage{multirow}
\usepackage{xcolor}
\usepackage{lineno}
\usepackage{appendix}
\usepackage{tabularx}
\usepackage{booktabs}
\usepackage{makecell}
\usepackage{geometry}
\usepackage{algorithmicx}
\usepackage{tabularx}
\usepackage{eurosym}
\usepackage{makecell}
\usepackage{multicol}
\usepackage{enumitem}
\usepackage{bm}
\newcommand{\nlabel}[1]{\textbf{\ensuremath{\bm{#1}}}}
\newcommand{\tlabel}[1]{\textbf{#1}}

\begin{document}
\begin{frontmatter}




\title{Peer-to-Peer Energy Trading in Dairy Farms using Multi-Agent Reinforcement Learning}


\author[inst1]{Mian Ibad Ali Shah\corref{cor1}}
\author[inst1]{Marcos Eduardo Cruz Victorio}
\author[inst2]{Maeve Duffy}
\author[inst1]{Enda Barrett}
\author[inst1]{Karl Mason}

\cortext[cor1]{Corresponding author. Email: m.shah7@universityofgalway.ie}
\cortext[]{Preprint of an article published in Applied Energy:
Shah, M.I.A., Victorio, M.E.C., Duffy, M., Barrett, E. and Mason, K. (2026). Peer-to-peer energy trading in dairy farms using multi-agent reinforcement learning. Applied Energy, 402, 127041. doi:10.1016/j.apenergy.2025.127041
Please cite the journal version.}

\affiliation[inst1]{%
  organization={School of Computer Science, College of Science and Engineering, University of Galway},
  addressline={University Road}, 
  city={Galway},
  postcode={H91TK33},
  country={Ireland}
}

\affiliation[inst2]{%
  organization={School of Engineering, College of Science and Engineering, University of Galway}, 
  addressline={University Road}, 
  city={Galway},
  postcode={H91TK33},
  country={Ireland}
}

\begin{abstract}


The integration of renewable energy resources in rural areas, such as dairy farming communities, enables decentralized energy management through Peer-to-Peer (P2P) energy trading. This research highlights the role of P2P trading in efficient energy distribution and its synergy with advanced optimization techniques. While traditional rule-based methods perform well under stable conditions, they struggle in dynamic environments. To address this, Multi-Agent Reinforcement Learning (MARL), specifically Proximal Policy Optimization (PPO) and Deep Q-Networks (DQN), is combined with community/distributed P2P trading mechanisms. By incorporating auction-based market clearing, a price advisor agent, and load and battery management, the approach achieves significant improvements. Results show that, compared to baseline models, DQN reduces electricity costs by 14.2\% in Ireland and 5.16\% in Finland, while increasing electricity revenue by 7.24\% and 12.73\%, respectively. PPO achieves the lowest peak hour demand, reducing it by 55.5\% in Ireland, while DQN reduces peak hour demand by 50.0\% in Ireland and 27.02\% in Finland. These improvements are attributed to both MARL algorithms and P2P energy trading, which together results in electricity cost and peak hour demand reduction, and increase electricity selling revenue. This study highlights the complementary strengths of DQN, PPO, and P2P trading in achieving efficient, adaptable, and sustainable energy management in rural communities.

\end{abstract}

\begin{keyword}
Dairy Farming \sep Energy Auction \sep Multi-agent Reinforcement Learning \sep Multi-agent Systems \sep P2P Energy Trading \sep    Renewable Energy
\PACS 0000 \sep 1111
\MSC 0000 \sep 1111
\end{keyword}

\end{frontmatter}


\newpage
\section*{Nomenclature}
\begin{multicols}{2}
\begin{description}[leftmargin=1.4em, labelsep=0.6em, itemsep=0.25ex, parsep=0ex, topsep=0.2ex]

\item[\textit{Sets and indices}]
\item[\nlabel{N}] Number of agents/farms in a community
\item[\nlabel{i, j, f}] Farm/agent index
\item[\nlabel{h}] Hour of day (0–23)
\item[\nlabel{d}] Day index (1–365)
\item[\nlabel{\beta}] Set of buyers in the double auction
\item[\nlabel{\sigma}] Set of sellers in the double auction
\\
\item[\textit{Parameters and constants}]
\item[\nlabel{\lambda_{\mathrm{buy}}(t)}] Utility grid time-of-use buying price at time $t$ (€/kWh)
\item[\nlabel{\lambda_{\mathrm{sell}}(t)}] Feed-in tariff selling price to grid at time $t$ (€/kWh)
\item[\nlabel{C_{\mathrm{bat}}}] Total battery energy capacity (kWh)
\item[\nlabel{C_{\mathrm{ch}}}] Maximum battery charge power (kW)
\item[\nlabel{C_{\mathrm{dis}}}] Maximum battery discharge power (kW)
\item[\nlabel{\mathrm{SoC}_{\min}}] Minimum allowable state of charge (\%)
\item[\nlabel{\mathrm{SoC}_{\max}}] Maximum allowable state of charge (\%)
\item[\nlabel{\alpha}] Learning rate (RL)
\item[\nlabel{\gamma}] Discount factor (RL)
\item[\nlabel{\Delta t}] Simulation time-step duration (h)
\\
\item[\textit{Energy variables}]
\item[\nlabel{L_f(t)}] Load demand of farm $f$ at time $t$ (kW)
\item[\nlabel{G_f(t)}] Total renewable generation of farm $f$ at time $t$ (kW)
\item[\nlabel{E_{\mathrm{pv},f}(t)}] PV generation of farm $f$ at time $t$ (kW)
\item[\nlabel{E_{\mathrm{w},f}(t)}] Wind generation of farm $f$ at time $t$ (kW)
\item[\nlabel{E_{\mathrm{tot},f}(t)}] Total available energy for farm $f$ (generation + schedulable) (kW)
\item[\nlabel{TBP(t)}] Total buying power in community at time $t$ (kWh)
\item[\nlabel{TSP(t)}] Total selling power in community at time $t$ (kWh)
\item[\nlabel{\mathrm{SoC}_f(t)}] State of charge of battery at farm $f$ (\%)
\item[\nlabel{B_{\mathrm{uc},f}}] Usable battery capacity at farm $f$ (kWh)
\item[\nlabel{B_{\mathrm{cc},f}(t)}] Current stored energy at farm $f$ (kWh)
\item[\nlabel{P_{\mathrm{ch},f}(t)}] Battery charging power (kW)
\item[\nlabel{P_{\mathrm{dis},f}(t)}] Battery discharging power (kW)
\item[\nlabel{Q_{\mathrm{ch},f}(t)}] Battery charge energy in step $t$ (kWh)
\item[\nlabel{Q_{\mathrm{dis},f}(t)}] Battery discharge energy in step $t$ (kWh)
\\
\item[\textit{Pricing and trading variables}]
\item[\nlabel{IBP(t)}] Internal buying price in P2P market at time $t$ (€/kWh)
\item[\nlabel{ISP(t)}] Internal selling price in P2P market at time $t$ (€/kWh)
\item[\nlabel{SDR(t)}] Supply–Demand Ratio $= TSP(t)/TBP(t)$
\item[\nlabel{P_{\beta,b}(t)}] Bid price offered by buyer $b$ (€/kWh)
\item[\nlabel{Q_{\beta,b}(t)}] Bid quantity by buyer $b$ (kWh)
\item[\nlabel{P_{\sigma,s}(t)}] Ask price offered by seller $s$ (€/kWh)
\item[\nlabel{Q_{\sigma,s}(t)}] Ask quantity by seller $s$ (kWh)
\item[\nlabel{O_b(t)}] Order book for buy orders at time $t$
\item[\nlabel{O_s(t)}] Order book for sell orders at time $t$
\\
\\
\item[\textit{Abbreviations}]
\item[\tlabel{AC}] Actor–Critic
\item[\tlabel{AI}] Artificial Intelligence
\item[\tlabel{CDA}] Continuous Double Auction
\item[\tlabel{DA}] Double Auction
\item[\tlabel{DQN}] Deep Q-Network
\item[\tlabel{DRL}] Deep Reinforcement Learning
\item[\tlabel{FiT}] Feed-in Tariff
\item[\tlabel{GA}] Genetic Algorithm
\item[\tlabel{IBP}] Internal Buying Price
\item[\tlabel{ISP}] Internal Selling Price
\item[\tlabel{KPI}] Key Performance Indicator
\item[\tlabel{LSTM}] Long Short-Term Memory
\item[\tlabel{MARL}] Multi-Agent Reinforcement Learning
\item[\tlabel{MAS}] Multi-Agent Systems
\item[\tlabel{NP}] Near-Peak (for model optimization)
\item[\tlabel{P2P}] Peer-to-Peer
\item[\tlabel{PPO}] Proximal Policy Optimization
\item[\tlabel{PSO}] Particle Swarm Optimization
\item[\tlabel{PV}] Photovoltaic
\item[\tlabel{RE}] Renewable Energy
\item[\tlabel{RL}] Reinforcement Learning
\item[\tlabel{SAM}] System Advisor Model
\item[\tlabel{SDR}] Supply–Demand Ratio
\item[\tlabel{SoC}] State of Charge
\item[\tlabel{TD}] Temporal Difference
\item[\tlabel{TFT}] Temporal Fusion Transformers
\item[\tlabel{ToU}] Time-of-Use

\end{description}
\end{multicols}

\section{Introduction}
\label{introduction}

The application of Artificial Intelligence (AI) to support dairy farms is experiencing rapid expansion, promising to transform farm management and operational efficiency \cite{de2023invited}. According to projections by Shine et al., global dairy consumption per capita is expected to rise by 19\% by 2050 \cite{shine2020energy}. However, the energy-intensive nature of milk production raises critical concerns about carbon dioxide ($CO_2$) emissions. Therefore, sustainable energy practices must accompany any increase in production to ensure the dairy sector’s long-term economic and environmental viability \cite{ben2017renewable}. Recent increases in energy costs, particularly during peak utility grid hours, have escalated electricity expenses for dairy farms \cite{upton2015investment}. This is because the timings of load demand for dairy farms are usually higher during these peak demand hours. Consequently, AI systems aimed at reducing carbon emissions and mitigating peak hour demand on the utility grid are increasingly sought after. Such systems could provide both environmental and financial benefits to dairy farms, ultimately supporting national economic growth.

A multi-agent system (MAS) is a collection of autonomous agents that work together toward a common objective, with each agent operating independently and communicating only with nearby peers \cite{adjerid2020multi, dorri2018multi}. Agents in MAS can function autonomously or semi-autonomously, making critical decisions based on local or global data to achieve design goals \cite{nair2018multi}. MAS effectively addresses complex problems using the divide-and-conquer approach, and has been widely applied in areas requiring distributed computing and control to create robust, flexible, and scalable systems \cite{shobole2021multiagent}. For microgrids, MAS offers a promising solution by providing a decentralized, fault-tolerant, and scalable control system \cite{elena2022multi}.

Although reinforcement learning (RL) algorithms have been utilized for sequential decision-making problems for decades, they traditionally struggle in high-dimensional environments. The reasons vary, one being the curse of dimensionality, where exponentially growing state and action spaces make exploration inefficient and learning computationally expensive. RL is also sample inefficient, as agents must visit states multiple times to estimate their value through trial and error, further slowing convergence. Sparse rewards and challenges in function approximation exacerbate these issues, leading to poor generalization and high resource demands. While techniques like dimensionality reduction, deep RL, and efficient exploration help, they do not fully resolve these challenges \cite{dulac2021challenges}. However, advancements in deep learning have enabled RL to produce optimal policies for sophisticated agents, allowing them to operate effectively in complex environments \cite{nguyen2020deep}. Multi-agent Deep Reinforcement Learning (MADRL) and Multi-agent Reinforcement Learning (MARL) are increasingly applied to peer-to-peer (P2P) energy trading within prosumer communities \cite{shah2024multi}.

The P2P energy trading model enables energy sharing within a microgrid of prosumers and consumers at an internal price, prior to transactions with external retailers to balance supply or demand deficits \cite{zhou2018evaluation}. P2P energy trading markets are classified primarily as centralized, decentralized, or community-based (distributed) markets. Centralized markets face challenges in computational and communication demands, as well as privacy concerns as the number of distributed energy resources increases \cite{umer2021novel}. In contrast, decentralized P2P markets allow prosumers and consumers to interact directly, establishing their own trading principles \cite{zhou2020state, guerrero2017study}. Hybrid P2P markets combine aspects of both centralized and decentralized systems, where a coordinator indirectly guides peers using price signals, providing a balance of privacy and autonomy. Research in P2P energy trading often leverages auction-based mechanisms \cite{zhou2020state, qiu2022mean}, such as the Double Auction (DA) model, which has also been adopted in this study. DA model is well suited for P2P energy trading due to its high transparency and efficiency \cite{qiu2022mean}\cite{malik2023comparative}.

While there is significant research on P2P energy trading and MARL, very few studies focus on rural agricultural settings, particularly dairy farms. Dairy farms have unique energy challenges: their electricity demand is substantial, highly variable, and often peaks during milking cycles that coincide with the most expensive utility tariff periods. These operational patterns, combined with rural grid constraints and limited access to ancillary services, set dairy farms apart from the residential or urban commercial settings that dominate existing research.

Our work addresses this gap by developing a MARL-based P2P trading framework specifically for rural dairy farms. Unlike previous studies, we adapt established algorithms (PPO/DQN) to handle the synchronized, high-load profiles and grid limitations typical of dairy operations. Additionally, we introduce two novel elements to this context: a price advisor agent and an auction-based market clearing mechanism. To our knowledge, this is the first time these components have been jointly applied to rural dairy farm energy trading. This tailored approach provides methodological novelty by addressing the specific operational and market challenges of rural dairy farms, which have been largely overlooked in prior work.

Hence, our research is particularly focused on solving the following challenges in the rural dairy farming sector as an application of P2P energy trading:

\begin{itemize}
    \item Highly cyclical and process-driven energy demand, with pronounced peaks during milking and cooling, resulting in inflexible load profiles.
    \item Significant and sudden surpluses or deficits due to the mismatch between generation and consumption, complicating local balancing within a P2P network.
    \item Limited potential for demand response, as many farm operations are time-sensitive and cannot be easily shifted to match renewable generation.
    \item Fewer but larger energy loads per user, leading to more pronounced swings in local demand and supply, and reduced aggregation benefits compared to urban settings.
\end{itemize}

This paper presents a method-centric contribution to privacy-preserving, market-aware MARL for P2P energy trading, designed for synchronized, inflexible loads in rural dairy communities, and demonstrates generalizability. All the following points are discussed in detail in Section \ref{results}.

\begin{enumerate}
    \item \textbf{Dairy-Constrained MARL Framework:} A novel RL scheduling problem is formulated in which agents optimize under endogenous, SDR-shaped prices within a double-auction market, while respecting critical dairy operational constraints (e.g., milking/cooling timing, must-run loads, pre-peak battery priming for SoC targets). The framework integrates dairy-aware mechanism design, including price-advisor tuning for synchronized peaks and battery management tailored to dairy duty cycles, ensuring realistic and robust energy management. The architecture is inherently portable, requiring only observation-vector and tariff/FiT adaptations for diverse agents.
    
    \item \textbf{Privacy-Preserving Policy-to-Market Interface:} A policy-to-market interface is operationalized to translate local MARL policies into feasible bid/ask schedules satisfying bounded-price rules. This interface improves community-level costs, peak demand, and revenues without compromising data privacy, since agents exchange only bids/asks and public prices.
    
    \item \textbf{Mechanism Attribution via Ablation:} Comprehensive ablation studies quantify the individual impact of the price advisor, pre-peak battery priming, and dairy-specific constraints. The analyses demonstrate how each component contributes to superior performance over rule-based and single-agent alternatives, highlighting the necessity of an integrated design for optimal behavior under dairy-specific conditions.
    
    \item \textbf{Demonstrated Generalizability and Scalability:}
    \begin{itemize}
        \item \textit{Heterogeneous Communities:} A companion study \cite{shah2025uncertainty} extends the MARL+price-advisor+DA architecture to heterogeneous dairy and residential communities with uncertainty forecasting, confirming effectiveness under agent diversity and forecast uncertainty, with comparable cost/peak reductions and revenue gains—demonstrating portability across prosumer mixes.
        \item \textit{Cross-Time/Space Transfer:} Robustness is validated through transfer tests training on Irish data and testing on Finnish data, demonstrating efficacy across different locations and seasonal regimes.
        \item \textit{Scalability Analysis:} Scalability is analyzed, showing stable market clearing times and graceful performance with increasing agent numbers, supporting practical deployment at community scale.
    \end{itemize}
    
    \item \textbf{Empirical Impact and Sector Relevance:} The framework delivers tangible benefits, including improved battery usage, reduced grid purchases during tariff peaks, consistent performance across seasonal variations, and significant cost reductions, peak charge mitigation, and revenue gains, with positive implications for the financial stability of rural prosumers.
\end{enumerate}

This research expands on previous work that implemented Q-learning for peer-to-peer energy trading in dairy farming. \cite{shah2024reinforcement}, \cite{shah2024peer}. In that study, Q-learning was implemented as a single agent along with rule-based agents to evaluate whether RL could improve simulator performance, which it successfully demonstrated. The Q-learning results of \cite{shah2024reinforcement} were compared to a rule-based approach \cite{shah2023multi} (which was also benchmarked against related literature), and the Q-learning ensemble method outperformed the rule-based approach. The results are summarized in Table \ref{irelandresult} of this paper. Here, a fully Multi-Agent Reinforcement Learning-based simulator is presented to model P2P energy trading, incorporate a price advisor, and utilize a Double Auction (DA) mechanism. The objectives of this research, aligned with the unique challenges of rural dairy farm energy management, are to:

\begin{enumerate}

    \item A privacy-preserving, market-aware MARL architecture that couples local value-based agents with a central price-advisor and a double auction. Agents optimize under endogenous, SDR-linked bounded prices (FiT/ToU), unlike typical RL-only schedulers.
    
    \item A practical policy-to-market interface that produces feasible bid/ask schedules, improving community-level costs, peaks, and revenues without sharing agent-internal information.

    \item Evaluate the impact of P2P energy trading using various RL models on peak hour (tariff) grid demand, farm electricity costs, and revenues from excess energy sales. This targets high tariff peaks, intermittent generation, and surplus/deficit management by quantifying cost reduction and revenue optimization.
    
    \item Benchmark the performance of purely MARL-based models against previous rule-based and rule-based with Q-learning ensemble models.This identifies the most adaptive and robust solutions for environments with limited flexibility and aggregation potential.
    
    \item Generalization evidence across locations, seasonal regimes, agent population sizes, and agent heterogeneity (dairy + residential) with uncertainty-aware forecasting; plus sensitivity analyses indicating robustness to battery/PV/load scale and agent count.
    
    \item Mechanistic insight via ablations removing the price advisor, removing battery pre-peak priming, and removing dairy constraints, attributing performance gains to each mechanism and explaining why the combined design outperforms rule-based and single-agent baselines.
 
\end{enumerate}

To achieve these objectives, two RL algorithms, Deep Q-Network (DQN) and Proximal Policy Optimization (PPO), are applied to manage load and battery storage for dairy farming agents. These models enable agents to make optimal decisions regarding energy transactions, including buying, selling, charging, and discharging battery storage, to facilitate efficient energy management within the dairy farming community.

\section{Background and Related Work}

In the field of sustainable energy distribution, peer-to-peer (P2P) trading within microgrids has emerged as a promising approach, promoting decentralization and efficiency. However, P2P energy trading faces challenges such as scalability, privacy, pricing mechanisms, and uncertainty. To address these issues, researchers have investigated community market mechanisms such as supply-demand ratios, midmarket rates, and bill-sharing strategies in P2P energy trading \cite{zhou2018evaluation}. Yet, these mechanisms apply the same cleared price to all participants, limiting self-interested traders from accessing trader-specific pricing signals \cite{long2017peer}. Alternatively, auction-based methods such as the double auction (DA) \cite{zheng2024multi} can determine individualized prices for each trader, increasing potential profit margins and creating a more competitive market. DA serves as an effective market mechanism, enabling energy transactions between multiple buyers and sellers while maintaining privacy and delivering overall economic benefits. Nonetheless, the inherent uncertainty and variability in trader behavior, DER generation, and load demand make DA-based markets highly dynamic in real-world scenarios, reducing their practicality. An approach that addresses many of the previously mentioned limitations—such as assumptions of prior knowledge (economic or otherwise), reliance on analytically tractable or differentiable objectives, computational complexity, agent interactions, data privacy, and the ability to adapt to changes in both agent behavior and the environment—is MARL \cite{zheng2022ai}. In essence, a RL agent learns to act through trial-and-error interactions with its environment \cite{may2023multi}. MARL, therefore, consists of multiple such agents that learn by interacting with each other within a shared environment \cite{johanson2022emergent}.

\subsection{Reinforcement Learning}

Reinforcement learning (RL) is a branch of machine learning that enables agents to make sequential decisions to maximize cumulative rewards within dynamic environments. Over recent years, RL has garnered substantial interest for complex applications like robotics, autonomous driving, and energy management \cite{arulkumaran2017brief}. RL agents interact with their environment to learn optimal actions that lead to favorable long-term outcomes rather than merely immediate rewards, utilizing mechanisms of exploration and exploitation to balance learning with performance \cite{mousavi2018deep}. The advent of deep reinforcement learning (DRL), which integrates RL with deep neural networks, has expanded the applicability of RL to high-dimensional state spaces, enhancing its utility in real-world applications such as autonomous energy management in smart grids \cite{franccois2018introduction}.

RL frameworks can operate in either single-agent or multi-agent contexts. In single-agent RL, agents work independently; however, in multi-agent reinforcement learning, multiple interacting agents seek to accomplish their objectives, which may involve cooperation or competition with others \cite{zhang2021multi}. This collaborative capacity of multi-agent systems allows for effective coordination strategies, such as task distribution and collective problem-solving, enabling each agent to contribute to the overarching goal while advancing individual objectives \cite{stone2010ad}. MAS utilize various mechanisms—including consensus algorithms, market-based systems, and auction-based negotiations—to facilitate agent interactions and enhance system performance. An example of MAS in action is in smart grid applications, where they manage decentralized resources like solar panels and batteries, optimizing energy distribution and improving resilience against system disruptions \cite{shobole2021multiagent}.

Recent advancements in MARL have further enhanced the adaptability of MAS, enabling them to learn optimal policies through continuous interaction and environmental feedback. MARL agents can learn strategies and coordination methods through trial and error, thus refining their actions over time without reliance on predefined programming \cite{zhang2021multi}.

\subsubsection{Q Learning}

Q-learning is a widely adopted off-policy reinforcement learning method and a key off-policy strategy. Since its introduction, extensive research has investigated its applications across various reinforcement learning and artificial intelligence challenges \cite{jang2019q}. In the realm of P2P energy trading, Q-learning serves as an effective tool for optimizing energy transactions and resource allocation among distributed agents. By leveraging past experiences and the rewards tied to specific actions, agents can formulate optimal strategies for energy buying, selling, and storage. This leads to more efficient use of renewable energy resources and grid infrastructure \cite{shah2024multi}. Through the balance of exploration and exploitation, agents can adapt to dynamic market conditions, maximizing trading profits while reducing reliance on centralized utility grids. Advanced Q-learning extensions, such as DQN and Prioritized Experience Replay, further improve scalability and performance in complex P2P energy trading scenarios. These advancements empower energy stakeholders to develop intelligent systems that optimize energy management, lower operational costs, and foster sustainability in decentralized energy markets.

In Q-learning, the action-value function \(Q(s,a)\) represents the expected cumulative reward for taking action \(a\) in state \(s\), and then following the optimal policy thereafter. The update rule for \(Q(s,a)\) is as follows:

\[
Q(s,a) \leftarrow (1 - \alpha) \cdot Q(s,a) + \alpha \cdot \left( r + \gamma \cdot \max_{a'} Q(s',a') \right)
\]

Where \( \alpha \) is the learning rate, \( r \) is the immediate reward, \( \gamma \) is the discount factor, and \( s' \) is the next state. The agent typically uses an \(\epsilon\)-greedy policy to balance exploration (trying new actions) and exploitation (choosing the best-known action).

The Q-learning algorithm can sometimes overestimate action values, even for suboptimal or incorrect actions. This overestimation arises from its reliance on the maximum predicted value during updates, which can inadvertently reinforce poor decisions. Recently, with the advancements in machine learning, several variants of Q-learning, such as Deep Q-learning, have emerged. Deep Q-learning combines traditional Q-learning with deep neural networks, enabling more practical and effective results, especially in high-dimensional and complex environments \cite{jang2019q}.

\subsubsection{Deep Q-Network (DQN)}

Traditional Q-learning relies on a Q-table to learn the optimal policy, but this approach struggles with large or continuous state and action spaces, as it demands significant memory and may fail to represent the policy effectively. This limitation is addressed by DQN, an off-policy reinforcement learning algorithm that uses deep neural networks to approximate Q-values, making it feasible to handle complex environments \cite{lv2019path}.

DQN builds upon traditional Q-learning by integrating deep learning to handle large state and action spaces. The central idea of DQN is to use deep neural networks to approximate Q-values, which represent the expected long-term rewards for taking specific actions in given states. This method enables the handling of complex, high-dimensional environments where traditional Q-learning approaches would struggle. In Q-learning, the value function is updated using the following temporal-difference (TD) formula:

\[
q(s_t, a_t) = q(s_t, a_t) + \alpha \left[ r_t + \gamma \max_{a_{t+1} \in A} q(s_{t+1}, a_{t+1}) - q(s_t, a_t) \right]
\]

where \(q(s_t, a_t)\) is the state-action value function at time \(t\), and \(r_t\) is the immediate reward. In DQN, a deep neural network is used to approximate the Q-values, and the network weights \(\theta_t\) are updated as follows:

\[
\theta_{t+1} = \theta_t + \alpha \left[ r_t + \gamma \max_{a_{t+1} \in A} q(s_{t+1}, a_{t+1}; \theta_{\text{TD}}) - q(s_t, a_t; \theta_t) \right] \nabla q(s_t, a_t; \theta_t)
\]

Here, \( \theta_{\text{TD}} \) is the target network, which is periodically updated to reduce training instability due to correlations in consecutive samples. The loss function is the difference between the TD target and the current Q-value, and the network is trained to minimize this loss using stochastic gradient descent (SGD).

\subsubsection{Proximal Policy Optimization (PPO)}

PPO is an on-policy deep reinforcement learning algorithm (Meng et al. in \cite{meng2023off} also proposed an off-policy PPO) that operates on the actor-critic (AC) architecture. In this setup, the Critic network estimates the value function, which provides an evaluation of state desirability, while the Actor network adjusts the policy by leveraging this value estimation to determine the best actions to take in various states \cite{gu2021proximal}. PPO is considered one of the most effective RL algorithms, delivering top-tier performance across a variety of complex tasks and benchmarks \cite{wang2020truly}.

PPO improves the stability and sample efficiency of policy gradient methods. Unlike traditional policy gradient methods, PPO aims to maximize expected rewards by directly optimizing the policy function. The strategy gradient method is used to maximize the expected reward, which is represented as:

\[
J(\theta) = \mathbb{E}_{\tau \sim \pi_{\theta}} \left[ \sum_{t=0}^{T} \gamma^t r_t \right]
\]

where \(\tau\) represents the state-action sequence and \(J(\theta)\) is the expected reward.

To avoid the instability caused by large policy updates, PPO introduces a clipping method that constrains the policy's change between old and new strategies. The loss function used in PPO is defined as:

\[
L_{\text{CLIP}}(\theta) = \mathbb{E}_t \left[ \min \left( \frac{\pi_{\theta}(a_t|s_t)}{\pi_{\theta_{\text{old}}}(a_t|s_t)} \hat{A}_t, \text{clip}\left( \frac{\pi_{\theta}(a_t|s_t)}{\pi_{\theta_{\text{old}}}(a_t|s_t)}, 1-\epsilon, 1+\epsilon \right) \hat{A}_t \right) \right]
\]

where \(\hat{A}_t\) is the advantage function, and \(\epsilon\) is the clipping threshold. This mechanism ensures that the policy updates remain within a reasonable range, helping to stabilize the training process.

By maintaining stable updates and preventing drastic policy changes, PPO enhances the efficiency and stability of reinforcement learning, particularly in complex and dynamic environments like autonomous navigation and industrial automation \cite{wang2024research}.

\subsection{P2P Markets and Related Work}

P2P energy trading, while still in its early stages \cite{wongthongtham2021blockchain}, has been explored through three types of architectures based on the degree and locus of coordination: centralized, decentralized (fully bilateral), and distributed/community P2P (often termed hybrid or platform-mediated P2P). In centralized P2P, an aggregator/community manager is the economic counterparty and schedules DERs; in decentralized P2P, peers bilaterally negotiate and settle without a coordinator; in distributed/community P2P, peers remain contractual counterparties while a neutral platform runs an automated market-clearing/matching service and feasibility checks. This work adopts the distributed/community P2P architecture. The coordinator does not buy/sell energy, hold inventory, or choose trading partners; it executes a fixed double-auction clearing rule and issues uniform clearing prices consistent with network feasibility. This framing follows the three-type taxonomy described in prior reviews and P2P design papers \cite{tushar2020peer, zhou2020state, qiu2022mean}.

Centralized approaches, such as PS-MADDPG \cite{qiu2021scalable} and non-cooperative game models \cite{zhang2019p2p}, focus on optimizing energy trading and pricing. Decentralized methods, including agent-based models \cite{guimaraes2021agent} and MARL frameworks \cite{zhou2020decentralized}, aim to maximize revenue and address factors like battery storage \cite{elkazaz2021hierarchical}. Distributed markets, which combine features of centralized and decentralized systems, emphasize scalability and autonomy by requiring minimal peer information and employing auction-based mechanisms like Continuous Double Auction (CDA) \cite{khorasany2018market, qiu2022mean}.

Research on distributed systems, particularly those relevant to this study, has largely focused on achieving financial benefits, scalability, and privacy. Distributed P2P energy-trading markets combine centralized and decentralized characteristics, where peer devices operate autonomously but are influenced by pricing signals from a central coordinator. This hybrid approach improves peer behavior while maintaining privacy and autonomy \cite{shah2024multi}. For example, in this study, only data related to load, generation, and pricing is shared with the auctioneer. Auction-based mechanisms are commonly employed in distributed P2P trading to establish prices, enabling buyers and sellers to compete until a mutually acceptable price is determined.

From a financial perspective, models like two-stage aggregated control \cite{long2018peer} and MARL-based frameworks \cite{pu2022peer} aim to optimize profits and cost savings. Scalability challenges are addressed through studies that focus on dynamic trading environments and large-scale participation \cite{qiu2022mean, charbonnier2022scalable}. Privacy and security concerns are mitigated using techniques such as DA-MADDPG \cite{qiu2021multi} and blockchain-based mechanisms \cite{abdelaziz2024blockchain}.

Rose et al. \cite{may2023multi} proposed a dynamic pricing strategy for renewable energy trading in microgrids to reduce costs and improve self-sufficiency. Prosumers with varying energy demand, supply, and battery capacities trade surplus energy at dynamic prices constrained by feed-in tariffs and consumer energy prices. A market-clearing mechanism balances demand and supply, while a social welfare function maximizes prosumer income. Using multi-agent reinforcement learning (PPO), the study demonstrated that dynamic pricing increased the proportion of households in profit (68\% vs. 66\% with fixed pricing) and encouraged strategic energy use, leading to better economic outcomes for the community. Similarly, Nat et al. \cite{uthayansuthi2024optimization} proposed a framework for optimizing P2P energy trading using clustering, forecasting, and deep reinforcement learning (DRL). K-means clustering with dynamic time warping (DTW) grouped energy patterns, while LSTM, DeepAR, and Temporal Fusion Transformers (TFT) handled 24-hour net load forecasting. PPO in a MARL setup enabled agents to trade energy in a model-based P2P double auction environment, reducing daily electricity costs for 300 customers while maintaining privacy and minimizing communication overhead \cite{uthayansuthi2024optimization}.

Recent advancements in P2P energy trading have also focused on enhancing trading robustness and efficiency through advanced optimization and AI techniques. For instance, Khodoomi et al. \cite{khodoomi2023robust} proposed a robust optimization framework for P2P energy trading, incorporating battery storage to address uncertainties in supply, demand, and pricing. While this approach improves trading reliability, it is primarily designed for generic energy systems and does not address the unique operational characteristics of agricultural environments. Similarly, Binyamin et al. \cite{binyamin2024artificial} explored the use of AI for energy community management in smart homes, demonstrating the potential of AI-driven coordination and optimization in residential renewable energy systems. However, applications in agriculture remain limited.

Recent studies have also investigated the application of DRL and hybrid optimization algorithms in energy management and load balancing. For example, Pradhan et al. \cite{pradhan2022intelligent} utilized deep reinforcement learning combined with parallel particle swarm optimization (PSO) for intelligent load balancing in cloud environments, highlighting the effectiveness of DRL in distributed, dynamic systems. Yousaf et al. \cite{yousaf2020comparative} provided a comparative analysis of various controller techniques, including genetic algorithms (GA) and PSO, for optimal control of smart nano-grids, emphasizing the importance of advanced optimization for efficient energy management. Additionally, Almahdi et al. \cite{almahdi2019constrained} demonstrated the potential of deep reinforcement learning for real-time energy management in microgrids, further supporting the applicability of learning-based methods in complex energy systems.

In contrast to prior work, this study applies multi-agent reinforcement learning (MARL) to peer-to-peer (P2P) energy trading among dairy farms, addressing the unique challenges of agricultural energy systems. The proposed MARL-based framework enables decentralized, adaptive decision-making suited to the dynamic and distributed nature of farm energy use. Unlike previous studies focused on residential, cloud, or microgrid settings, this research targets agricultural-specific issues such as variable production, unique consumption profiles, and the need for scalable coordination among multiple farms. The algorithm is designed to optimize trading strategies during grid peak hours, when tariffs are highest, reducing energy imports and costs, which is distinct from simply managing the farm’s own peak demand. By incorporating financial benefits, privacy, scalability, user satisfaction, load management, and a transparent auction mechanism, this approach offers a comprehensive solution for P2P energy trading in agriculture. Integrating MARL with P2P trading in this context fills a critical gap in the literature and lays the groundwork for intelligent energy management in farming communities.

\section{Methodology}
\label{methodology}

This study presents a MARL-based algorithm to enable distributed P2P energy trading among dairy farms. The proposed approach utilizes MARL to allow farms to sell excess energy generated from renewable energy resources to nearby farms. Each farm agent learns to manage its load and battery through the MARL algorithm. Once trained, the agents decide when to sell or purchase electricity and when to utilize their batteries based on factors such as energy demand and supply, State of Charge (SoC), market prices, and economic considerations (e.g., trading vs. storing). Together, these agents collaborate to meet the system's energy demands efficiently, minimizing electricity purchase costs and grid dependence while maximizing revenue from electricity sales.

The system operates on a distributed P2P energy trading model, where only data related to energy generation and load is shared with the auctioneer. However, market clearance is conducted in a centralized manner. Similar to a decentralized market, each participant independently manages their own load, generation, and battery usage without revealing any additional information to external entities. This approach ensures privacy while maintaining efficient market operations.

Participants share their excess or deficit electricity data with a centralized coordinator, which acts as both an auctioneer and advisor for the community. The advisor is a non-transacting, non-discretionary clearing service. Specifically: (i) it is never a counterparty in any matched trade; (ii) it does not aggregate energy or hold inventory; (iii) it implements a fixed double-auction clearing rule to determine price and allocations subject to feasibility; and (iv) it does not select trading partners beyond the algorithmic matching implied by submitted bids and asks. Peers retain full independence to set and modify their own bids and asks. Contractual counterparties and settlements are between peers. All cleared trades are peer-to-peer; the platform provides registry, metering reconciliation, and automated clearing computations only. This advisor assesses market conditions, such as current supply and demand for electricity, and calculates an Internal Selling Price (ISP) and Internal Buying Price (IBP) using the Supply and Demand Ratio (SDR) technique. The SDR method allows real-time price determination based on the system's energy demand and supply \cite{liu2017energy}. The ISP represents the price at which a participant sells surplus energy directly to another participant, ensuring fair energy exchange between producers and consumers. Similarly, the IBP reflects the price at which a participant purchases energy from another participant, indicating the cost incurred by the buyer to acquire energy within the P2P platform. These prices promote efficient energy trading by ensuring fair compensation and cost transparency.

Using the ISP and IBP, each farm agent can submit bids or offers specifying the price and quantity of electricity they wish to trade. The auctioneer agent processes these bids using a double auction algorithm to clear the market, as described in detail in the appendix.

If any excess electricity remains after market clearing, it is sold to the grid; conversely, if there is a shortage, electricity is purchased from the grid. This mechanism aims to reduce the dependency on the power grid and improve the self-sufficiency of energy for farms.

\begin{figure*}
    \centering
    \includegraphics[width=1\textwidth]{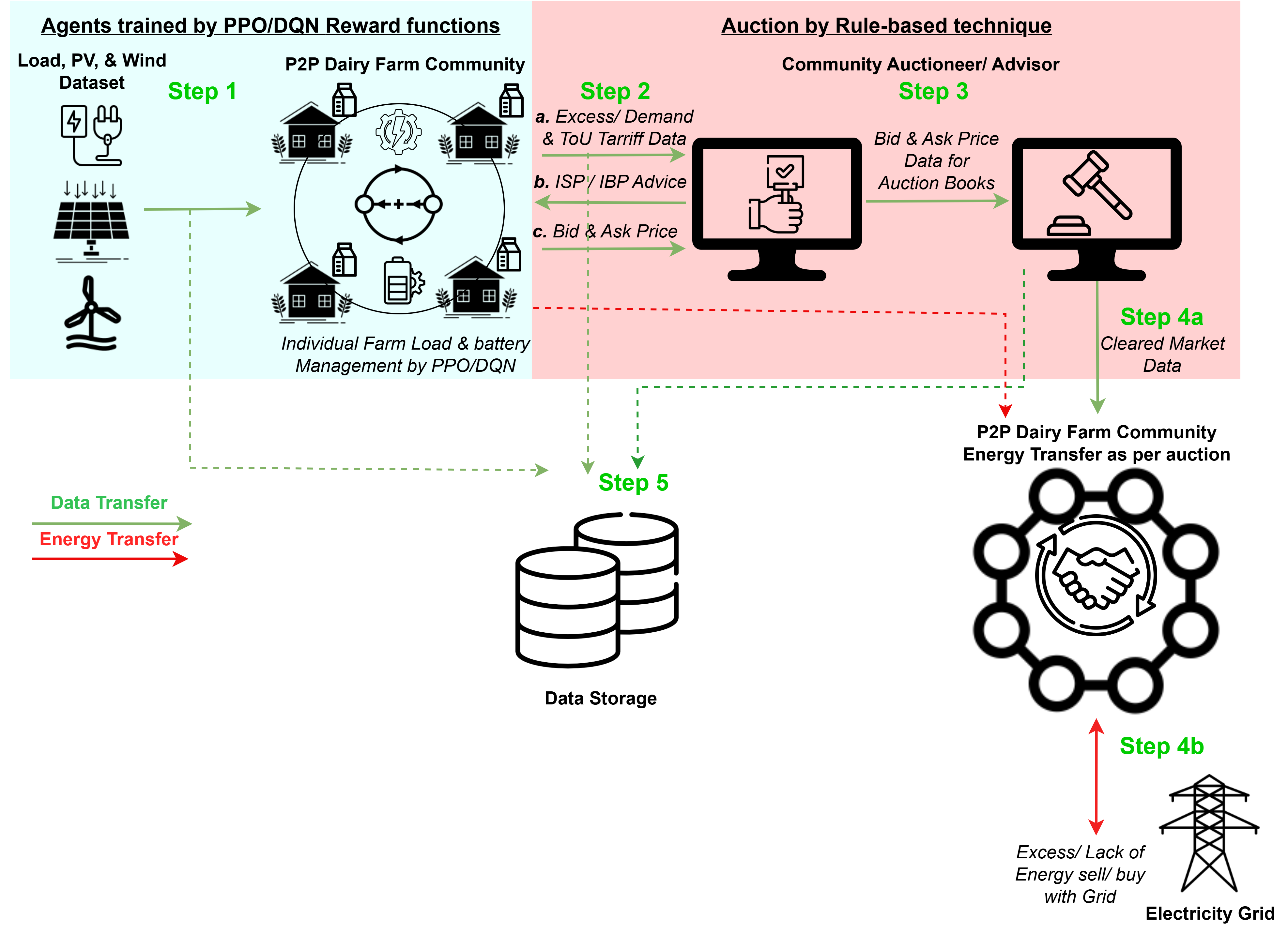}
    \caption{Process flow of MARL MAPDES simulator}
    \label{methodology figure}
\end{figure*}

\begin{figure}
    \centering
    \includegraphics[width=0.75\textwidth]{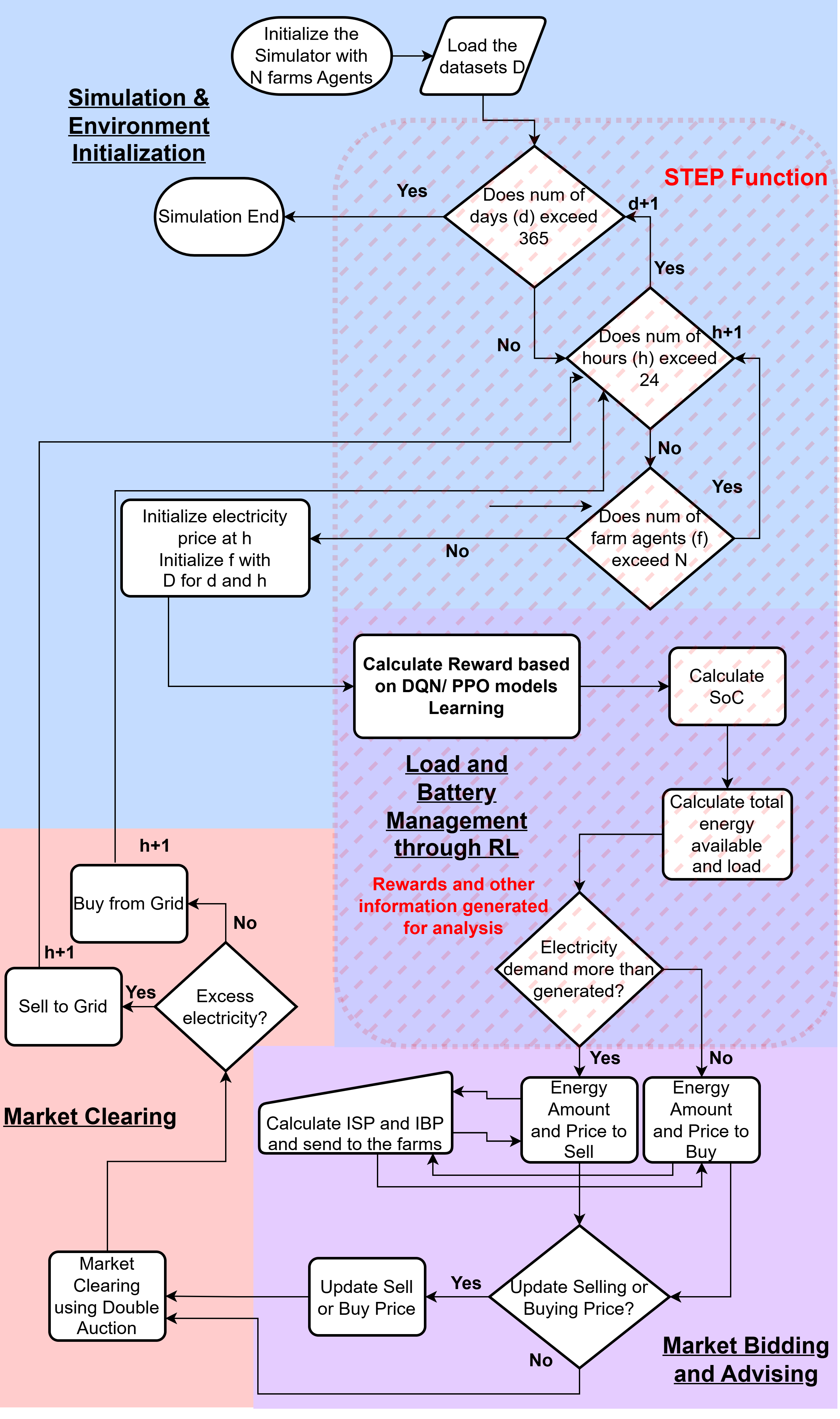}
    \caption{Flowchart of the simulator}
    \label{flowchart}
\end{figure}

The novel aspects of this work compared to our prior Q-learning studies are: (a) a full multi-agent RL implementation; (b) implementation and comparison of two reinforcement learning algorithms, DQN and PPO, to identify the best-performing approach; and (c) benchmarking against multiple baselines. The paper compares three versions of the MAPDES simulator: Rule-based, Q-learning and Rule-based Ensemble, and MARL, with the primary focus on the MARL approach, which is the most recent and central to this study. The process flow of the MARL simulator is illustrated in Figure~\ref{methodology figure}. The illustrations for the other two simulators are available in the published works~\cite{shah2024peer, shah2024reinforcement}. All relevant equations (rules) from these publications are included in~\ref{loadandbatterymanagement}, and the corresponding pseudocode is provided in Algorithm~\ref{RBpseudocode}. A stepwise flowchart of the MARL simulator is also shown in Figure~\ref{flowchart}, with pseudocode in Algorithm \ref{MARLpseudocode}, outlining the techniques and equations used in this study.

\begin{algorithm}
\caption{Rule-based MAPDES Model Pseudocode from \cite{shah2024peer}}
\label{RBpseudocode}
\begin{algorithmic}[1]
    \State Initialize simulator (load dataset $D$)
    
    \For{each day $d$ in year}
        \For{each hour $h$ in day}
            \For{each farm $f$}
                \State Check for RE resources availability in $f$
                \State Calculate electricity price $\lambda_{\text{buy}}$ based on time period (night/day/peak) (Eqn \ref{lambdabuy})
                \State Initialize $f$ with values from $D$ for $d$ and $h$
                \State Calculate usable battery capacity $B_{\text{uc}}$ based on current capacity and max discharge (Eqn \ref{usbalecapacity})
                \State Calculate total energy generation $E_{\text{tot}}$ by summing PV, wind, and battery (Eqn \ref{totalgeneration})
                \State Get farm load $E_l$ from $D$
                \State Update battery SoC based on charging/discharging status (Eqn \ref{batterypercent})
                \State Make battery charge/discharge decision based on:
                      \begin{itemize}
                        \item Total energy vs load
                        \item Current SoC
                        \item Electricity price
                        \item Using Eqns \ref{equ4}, \ref{equ5}, \ref{equ7}
                      \end{itemize}
                \State Calculate battery charging capacity and discharge percentage (Eqns \ref{chargingpower}, \ref{chargingpercentage})
                \State Determine energy trading actions:
                      \begin{itemize}
                        \item Calculate excess energy to sell (Eqn \ref{excessenergy})
                        \item Calculate energy to buy (Eqn \ref{energybuy})
                        \item Based on RE availability, battery status, and prices
                      \end{itemize}
            \EndFor
            \State Calculate P2P selling/buying prices using supply-demand ratio (Eqns \ref{sdr}, \ref{ISP}, \ref{IBP})
            \State Match P2P buyers and sellers (Algorithm \ref{auction})
            \State Purchase/sell additional/excess energy from/to grid (Algorithm \ref{auction})
        \EndFor
    \EndFor
\end{algorithmic}
\end{algorithm}

\subsection{Datasets and Infrastructure}
\label{dataset&infra}

This study incorporates multiple case studies, detailed in the results section. Simulations are based on datasets from Finland and Ireland, each involving 10 farms. The Finnish farm load data is sourced from Uski et al. \cite{uski2018microgrid}, while the Irish farm load data is generated using a dairy farm load modeling algorithm by Khaleghy et al. \cite{khaleghy2023modelling}. PV generation profiles for both locations are produced using the System Advisor Model (SAM) \cite{samNrel}, based on the geographic coordinates of Helsinki and County Dublin. All datasets are uniformly sampled at one-hour intervals over a full year (8,760 hours). Distinct years are used for training and testing, and the stochasticity in agent actions and market dynamics ensures sufficient variability across episodes.

In the practical implementation phase, where the RL framework is deployed in real-world conditions separate from the learning or evaluation phases, data sampling may be adjusted to enhance accuracy. This adjustment involves integrating real-time data from additional equipment, such as smart meters, sensors for renewable energy generators, and historical consumption records, to better reflect operational conditions. These improvements will help evaluate the quality and reliability of the data sources. Furthermore, data validation and cleaning procedures will be applied to detect and address outliers, correct errors, and interpolate missing data points, ensuring the datasets are robust and reliable for practical use.

Table \ref{FarmsSpecs} summarizes the farm sizes and their PV capacities for Ireland and Finland. Farms range in herd size from 30 to 70 cows, with PV capacities adjusted for each region. For example, farms with a herd size of 30 have 10 kW of PV in Ireland and 20 kW in Finland, while farms with a herd size of 70 have 20 kW in Ireland and 50 kW in Finland.

Based on the literature \cite{SEAI2024}, the renewable generation capacity typically installed is about 40\% of the total annual load of the farms. The PV capacities differ between Ireland and Finland due to variations in energy demand throughout the year. To reflect this, PV capacities have been set to approximately 40–50\% of the farms' total load, ensuring they align with regional energy needs. Battery storage is incorporated into the simulation using the Tesla PowerWall \cite{Teslapower}, with each farm utilizing one battery unit. Key specifications for the batteries are outlined in Table \ref{batteryspecs}.

The infrastructure designed for this study is scalable, supporting a variable number of farms beyond the quantities used in the case studies. All the farms are considered as prosumers and equipped with appropriate PV and batteries.

\begin{table}[h]
\centering
\begin{tabular}{lccc}
\hline
\makecell{\textbf{Farm} \\ } & \makecell{\textbf{Farm (Herd)}\\\textbf{Size}} & \makecell{\textbf{Ireland PV}\\\textbf{(kW)}} & \makecell{\textbf{Finland PV}\\\textbf{(kW)}} \\ \hline
Farm 1,2 & 30 & 10 & 20 \\
Farm 3,4 & 40 & 10 & 30 \\
Farm 5,6 & 50 & 20 & 40 \\
Farm 7,8 & 60 & 20 & 50 \\
Farm 9,10 & 70 & 20 & 50 \\ \hline
\end{tabular}
\caption{Farm Size and RE Specifications for both Ireland and Finland Datasets}
\label{FarmsSpecs}
\end{table}

\begin{table}[ht]
    \centering
    \caption{Battery Specifications of Tesla PowerWall \cite{Teslapower}}
    \begin{tabular}{ll}
        \hline
        \textbf{Parameter} & \textbf{Value} \\
        \hline
        Total Capacity & 14 kWh \\
        Usable Capacity & 13.5 kWh \\
        Real/ max continuous power & 5 kW (charging \& discharging) \\
        \hline
    \end{tabular}
    \label{batteryspecs}
\end{table}

\subsection{Model Design}
\label{techniques}

The proposed MARL MAPDES is a Python-based simulator implemented using a MARL framework, where each agent represents a farm participating in an energy trading system. The core of this environment is built upon the PettingZoo library, which facilitates the creation of parallel environments for multiple agents. Each agent is designed to simulate a dairy farm with specific energy generation and consumption profiles, where agents must decide on their energy consumption and production strategies while optimizing their rewards based on market dynamics and battery usage.
The simulator operates through three key stages to enable efficient energy trading and management within a P2P network. First, individual farm agents use RL algorithms like Proximal Policy Optimization (PPO) or Deep Q-Network (DQN) for load and battery management. Through these algorithms, each agent learns optimal strategies for consuming, storing, or selling energy based on their specific needs. Next, the community’s ISP and IBP are calculated by the auctioneer, which determines the price at which farms sell excess energy and purchase energy. These prices facilitate fair energy exchanges within the network. Finally, the auctioneer clears the market using a DA mechanism, matching buyers and sellers based on their bids and ensuring efficient energy distribution. The simulator is flexible, allowing it to run for durations ranging from an hour to several years, providing comprehensive insights into the grid's load dynamics and potential profits or losses from energy trading within the P2P network. The following sections of Load and Battery Management, Electricity Pricing, and Market Clearing and Energy Trading has been discussed in detail in published work \cite{shah2024peer}, hence all the required equations are added to \ref{loadandbatterymanagement} of this paper and the topics are discussed briefly as the proposed model uses the same concepts.

\subsection{RL Environment Setup}

Both models (PPO and DQN) were trained over 2.5 million episodes to learn the optimal actions for each agent. Each episode represents a complete year, spanning 8760 hours. The observation space for each agent is a four-dimensional vector comprising energy demand, generation, battery state-of-charge (SoC), and time of day. The discrete action space includes eight possible actions, such as charging, discharging, or trading energy with the grid. Grid buy and sell prices fluctuate dynamically based on the time of day and peak demand, influencing agents' decisions. The agent class inherits from ParallelEnv in PettingZoo, simulating the energy trading environment. Training is conducted using the Stable-Baselines3 RL library \cite{stablebaselines3docs} for both DQN and PPO.

Each agent represents a farm with a unique identifier mapped through a dictionary to access energy load and generation data. Agents are trained on yearly hourly energy load and generation profiles from Ireland and Finland, which dynamically update to reflect real-time conditions. Each agent is equipped with a battery with defined charging and discharging limits, while grid prices vary by time, with higher tariffs during peak hours as described in Equation \ref{lambdabuy}. Agents select from eight discrete actions, such as charging, buying, selling, discharging, or using self-generated energy, based on their state. Further details of the agent setup are provided in \ref{agentsetup}.

\subsubsection{Load and Battery Management}

Battery usage and renewable energy management are based on the equations provided in \ref{loadandbatterymanagement}, which were utilized in previously published work to calculate usable battery capacity based on current and maximum discharge capabilities, ensuring only available charge is used. These equations served as the foundation for formulating the reward functions for managing battery and renewable energy resources, as detailed in Section \ref{rewardfunctions}.

The model prioritizes battery charging when renewable energy generation exceeds farm load and the battery level is below 90\%, preventing overheating and extending battery life. Decisions on battery discharge and energy purchases depend on renewable generation, battery state, and energy prices. When renewable energy is unavailable and the battery falls below 10\%, energy is purchased from the grid, considering off-peak and peak pricing. If renewable energy is available but no battery is installed, the model evaluates whether to buy or sell energy based on generation and load.

After meeting the farm’s load requirements, any surplus energy is available for sale, with SoC factored in. Energy purchase calculations consider renewable availability, battery status, energy prices, and demand, optimizing energy utilization and reducing costs. Once individual load and battery management are in place on farms, transactions proceed to either sell excess energy or purchase additional energy from the community or grid.

\subsection{RL Reward Functions}
\label{rewardfunctions}

The reward function for the RL models (PPO and DQN) in this study is designed based on the load and battery management equations published in previous work \cite{shah2024peer}. The formulation of the reward function helps each agent make optimal energy management decisions by incentivizing self-sufficiency, minimizing grid dependency, and maximizing efficient use of available resources. The function assigns rewards based on actions the agent takes concerning energy generation, load, SoC, and grid conditions. Through this multi-action, condition-based reward scheme, agents learn to balance their battery management and energy transactions with other agents. The following is a detailed breakdown of each possible action and its corresponding reward calculation.

These reward functions represent the optimal configurations identified through extensive testing, experimentation, and the application of domain expertise. The following terms are used throughout the reward functions:

\begin{itemize}
    \item $C_{\text{bat}}$ : Total battery capacity(kWh), $C_{\text{charge}}$: Battery charging capacity (kWh), $Q$: Discharge amount (kWh)
    \item SoC: State of Charge of the battery (\%)
    \item $G$: Generation (kW), $L$: Load (kW)
    \item $T_{\text{grid}}$: Grid tariff, which can be:
        $N$ (Night), $NP$ (Near Peak for optimized battery management), $P$ (Peak), or $D$ (Day)
\end{itemize}

The reward functions are defined as follows:

1. \textbf{Charge and Buy}  
   The agent purchases energy to charge its battery when either the SoC is low ($\text{SoC} \leq 50\%$) and $G$ is insufficient to meet $L$, or during $T_{\text{grid}} \in \{N, NP\}$ with low SoC. It is worth noting that NP is not a tariff from the grid; rather, it is an optimization variable used in the model to keep the batteries charged before the grid peak hours begin. The reward function is:
   \[
    \text{reward} = 
    \begin{cases} 
    1, & \text{if } (\text{SoC} \leq 50\% \text{ AND } G < L) \text{ OR } \\
    & (\text{SoC} \leq 50\% \text{ AND } T_{\text{grid}} \in \{N, NP\}) \\
    0, & \text{otherwise}
    \end{cases}
    \]
   The SoC is updated as: $\text{SoC} = \min(\text{SoC} + \frac{C_{\text{charge}}}{C_{\text{bat}}} \times 100, 100)$.\\

2. \textbf{BUY}  
   The agent buys energy when $G$ is insufficient and SoC is critically low. The reward is:
   \[
   \text{reward} = 
   \begin{cases} 
      1, & \text{if } G < L \text{ AND SoC} < 10\% \text{ AND } T_{\text{grid}} \neq N \\
      0, & \text{otherwise} 
   \end{cases}
   \]

3. \textbf{SELL}  
   The agent sells excess energy when SoC is sufficiently charged. The reward is:
   \[
   \text{reward} = 
   \begin{cases} 
      1, & \text{if } G > L \text{ AND } (\text{SoC} \geq 90\% \text{ OR } \\
      & (\text{SoC} \geq 20\% \text{ AND } T_{\text{grid}} = P)) \\
      0, & \text{otherwise} 
   \end{cases}
   \]

4. \textbf{Discharge and Sell}  
   The agent discharges and sells energy when there's excess $G$ and sufficient SoC. The discharge amount is:
   \[
   Q = \min(C_{\text{charge}}, \frac{\text{SoC} \times C_{\text{bat}}}{100})
   \]
   The reward function is:
   \[
    \text{reward} = 
    \begin{cases} 
    1, & \text{if } G > L \text{ AND } ((\text{SoC} \geq 20\% \text{ AND } T_{\text{grid}} = P) \text{ OR } \\
    & (\text{SoC} \geq 90\% \text{ AND } T_{\text{grid}} \neq P)) \text{ AND } E > 0 \\
    0, & \text{otherwise}
    \end{cases}
    \]
   where $E$ is the excess generation after discharge.

5. \textbf{Discharge and Buy}  
   When $G$ is insufficient, the agent can use battery power to reduce grid purchases. The reward is:
   \[
    \text{reward} = 
    \begin{cases} 
    1, & \text{if } G < L \text{ AND SoC} \geq 10\% \text{ AND } Q > 0 \\
    0, & \text{otherwise}
    \end{cases}
    \]
   where $Q = \min(C_{\text{charge}}, L - G)$ is the discharge amount.\\

6. \textbf{SELF}  
   The agent operates in self-sufficient mode when $G$ matches closely $L$ (difference assumed less than 0.1 kW):
   \[
    \text{reward} = 
    \begin{cases} 
    1, & \text{if } |G - L| \leq 0.1 \\
    0, & \text{otherwise}
    \end{cases}
    \]

7. \textbf{Self and Charge}  
    The agent charges the battery with excess $G$ during non-peak hours. Initially, it is verified whether charging is required and if the required charge is less than or equal to the remaining capacity of the battery, as:
   \[
   Q_{\text{charge}} = \min(C_{\text{charge}}, G - L)
   \]
   \[
    \text{reward} = 
    \begin{cases} 
    1, & \text{if } Q_{\text{charge}} > 0 \text{ AND } G > L \text{ AND SoC} \leq 80\% \text{ AND } T_{\text{grid}} \neq P \\
    0, & \text{otherwise}
    \end{cases}
    \]

8. \textbf{Self and Discharge}  
   The agent discharges the battery to supplement self-consumption:
   \[
    \text{reward} = 
    \begin{cases} 
    1, & \text{if } G < L \text{ AND SoC} \geq 20\% \text{ AND } |Q - (L-G)| \leq 0.1 \\
    0, & \text{otherwise}
    \end{cases}
    \]

The discrete reward system provides clear feedback to the agent about whether its actions are desirable (\textit{reward} = 1) or not (\textit{reward} = 0). The conditions for each action have been refined based on practical thresholds for SoC, $T_\mathrm{grid}$, and the energy balance between $G$ and $L$. These modifications reflect the actual implementation in the Python code and provide more precise control over the agent’s behavior in different scenarios. This learning process and the agent’s decision-making pattern will be clearly visible in the reward convergence graphs, which will be discussed in the results section of the paper. These graphs will show how the model progressively adjusts its behavior to maximize the reward over time. These reward structures not only incentivize actions aligned with the broader goals of the P2P energy network but also update the agent’s battery state based on each action. The structure helps agents manage both their energy needs and surpluses, which, in turn, informs the setting of the ISP and IBP in the centralized market clearing process. This function effectively aligns each agent’s individual rewards with the community’s goals of grid independence and efficient energy trading. 

To ensure reproducibility and formal clarity, the state transition functions governing agent-environment interactions have been explicitly defined, and the detailed pseudocode is provided in \ref{statestransition}.

The eight discrete action choices for each agent are designed to cover a comprehensive range of energy management strategies within the P2P trading environment. Each action is defined with specific preconditions related to SoC, current grid tariff, and the agent's energy balance (generation vs.\ load). While this action set aims for broad coverage, it is acknowledged that some actions may exhibit conditional redundancy or infeasibility depending on the current state of the environment. For instance, an action like \texttt{CHARGE\_AND\_BUY} might effectively reduce to a simple \texttt{BUY} if the battery is already full, or if charging conditions are not met. Similarly, \texttt{DISCHARGE\_AND\_SELL} or \texttt{DISCHARGE\_AND\_BUY} actions are only effective when the battery has sufficient charge and specific energy balance conditions are present. The \texttt{SELF} action, intended for near-perfect self-consumption, may also overlap with other actions if there is only a slight imbalance. This design choice prioritizes providing the agent with a rich set of options to explore, allowing the reinforcement learning algorithm to discover optimal policies by navigating these conditional validities.

\subsubsection{Electricity Pricing}

To facilitate energy trading among prosumers, an appropriate price for each hour is calculated based on the community’s total load and generation. The internal buying and selling prices (ISB/ ISP) are determined using the SDR technique, as proposed by Liu et al. in \cite{liu2017energy}. Prosumers have varying loads and energy generation, and their net power (generation minus consumption) determines whether they buy or sell electricity. The Total Selling Power (TSP) and Total Buying Power (TBP) are computed for the community, with the net power determined as the difference between TBP and TSP. Internal prices are set considering the feed-in tariff (FiT), the utility grid’s electricity price, and economic balance principles.

The relationship between price and SDR is inverse-proportional, with a piecewise function for the ISP, as shown in Equation \ref{sdr}. When SDR = 0, indicating there is no selling power available within the community, energy is procured from the utility grid at the $\lambda_{\text{buy}}$ price. If SDR $\geq$ 1, excess energy is sold at the $\lambda_{\text{sell}}$ (FiT) price. When SDR is between 0 and 1, the selling price fluctuates between $\lambda_{\text{sell}}$ and $\lambda_{\text{buy}}$.

For instance, if TBP $>$ TSP, indicating high demand, the buying price is high; if TSP $>$ TBP, the buying price is low. The IBP and ISP are based on SDR, where the missing power is bought at the $\lambda_{\text{buy}}$ price. Prosumers adjust their power consumption based on SDR to benefit from the price fluctuations.

\begin{equation} \label{sdr} SDR = \frac{TSP}{TBP} \end{equation}

\begin{equation} \label{ISP} \text{ISP} = \begin{cases} \frac{\lambda_{sell} \times \lambda_{buy}}{(\lambda_{buy} - \lambda_{sell}) \times \text{SDR} + \lambda_{sell}}, & 0 \leq \text{SDR} \leq 1 \\ \lambda_{sell}, & \text{SDR} > 1 \end{cases} \end{equation}

\begin{equation} \label{IBP} \text{IBP} = \begin{cases} \text{ISP} \times \text{SDR} + \lambda_{buy} \times \text{(1-SDR)}, & 0 \leq \text{SDR} \leq 1 \\ \lambda_{buy}, & \text{SDR} > 1 \end{cases} \end{equation}

As an example, if the community’s TBP is higher than the TSP, and the SDR is less than 1, the ISP and IBP are calculated as 0.39 €/kWh and 0.52 €/kWh, respectively, with higher prices for prosumers selling electricity. If the TSP is higher than the TBP, SDR $>$ 1, and the ISP equals the FiT (0.135 €/kWh), while IBP equals the grid price (0.66 €/kWh), enabling prosumers to sell electricity at the FiT and purchase it at a lower price. These price fluctuations allow both buyers and sellers to benefit from the market clearing.

\subsubsection{Market Clearing and Energy Trading}

During the final simulation phase, data from earlier stages is leveraged to enable market clearance, supporting energy trading among community members and the utility grid in alignment with the cleared market outcomes. This study utilizes a DA auction mechanism, inspired by Qiu et al.~\cite{qiu2022mean, qiu2021multi}. The auction algorithm is adapted for dairy farms, with the auctioneer requiring only load, generation, and pricing data from participants, ensuring data privacy.

The DA market matches buyers and sellers based on submitted bids/offers, controlled by participants. The auction operates with hourly resolution, and the auctioneer clears the market, publishing public outcomes (prices and quantities) at the end of each period~\cite{qiu2022mean}.

\textbf{Double Auction Mechanism}\\
The auction system consists of buyers, $\beta$, and sellers, $\sigma$. Each buyer \textit{b} specifies a price $P_{\beta,b}$ and quantity $Q_{\beta,b}$ they wish to purchase, while each seller \textit{s} specifies their selling price $P_{\sigma,s}$ and quantity $Q_{\sigma,s}$ of energy. The auctioneer manages two order books, $O_b$ for buy orders and $O_s$ for sell orders, sorted by price.

\textbf{Order Booking}\\
The auctioneer manages bid and ask order books. Bids are sorted in descending order of price, while asks are sorted in ascending order. Orders, including trading price and energy quantity, are added to the respective books. The prices must fall between the FiT and ToU tariff rates.

\begin{algorithm}
\caption{MARL MAPDES P2P Energy Trading Simulation}
\label{MARLpseudocode}
\begin{algorithmic}

\State \textbf{Initialize Environment:}
    \State Set number of farm agents $N$, dataset $D$, day counter $d = 0$, hour counter $h = 0$.

\While{$d \leq 365$} \Comment{1-year simulation}
    \If{$h > 24$} \Comment{Reset daily}
        \State $d \gets d + 1$, $h \gets 0$
    \EndIf
    \If{$d > 365$} \textbf{End Simulation} and \textbf{Exit loop} \EndIf

    \For{each farm agent $f$ in $(1, N)$}
        \State Initialize electricity price for hour $h$.
        \State Initialize agent $f$ with $D$ for day $d$, hour $h$.
        \State Calculate reward based on DQN/PPO model.
        \State \textbf{Load and Battery Management:}
            \State Calculate battery SoC, total energy, and demand.
            \If{demand $>$ generated energy} Set energy amount and price to \textbf{buy}.
            \Else \ Set energy amount and price to \textbf{sell}.
            \EndIf

        \If{excess electricity available} Execute \textbf{Add to Auction Seller Book}.
        \Else \ Execute \textbf{Add to Auction Buyer Book}.
        \EndIf
    \EndFor

    \State \textbf{Market Clearing Stage:}
        \State Calculate ISP and IBP, send to farm agents.
        \For{each farm agent}
            \If{sell/buy price update needed} Update price based on market.
            \EndIf
        \EndFor
        \State Perform Market Clearing using \textbf{Double Auction}.
    \State $h \gets h + 1$ \Comment{Increment hour}
\EndWhile

\State \textbf{End Algorithm}
\end{algorithmic}
\end{algorithm}

\textbf{Market Clearing}\\
The matching algorithm pairs buy and sell orders sequentially, determining the market clearing price using the mid-pricing method. The transaction quantity is the minimum of the matched orders. Any remaining orders or energy quantities are settled with the utility company at ToU or FiT prices. FiT compensates RE producers for feeding electricity into the grid, while ToU prices vary based on time of day. Traders' pricing strategies are constrained within the FiT and ToU limits to ensure economic balance.

The auction and market clearing algorithm implemented in this simulator is presented in Algorithm~3 in Appendix~B. For full methodological transparency, we clarify that the detailed workings of the double auction mechanism, including the matching strategy, price priority, and supply-demand balancing, are based on the approaches described in the parent papers by Qiu et al.~\cite{qiu2022mean, qiu2021multi}, where the mechanism is explained in depth.

The auction and market clearing algorithm of the simulator is presented in Algorithm~\ref{auction} in the appendix. The complete pseudocode for the MARL simulator is presented in Algorithm~\ref{MARLpseudocode}, detailing the steps of data initialization, reward generation through model learning, electricity pricing advisement, auction, and energy trading, as discussed in Section~\ref{techniques}.

\section{Experimental Results}
\label{results}

To evaluate the efficacy of the MARL simulation for energy trading among prosumer farms, a year-long simulation with a one-hour time step has been conducted, comparing the performance of different approaches, rule-based as baseline model (explained in \ref{loadandbatterymanagement}), rule-based + Q-learning, and RL models (DQN and PPO) with and without P2P trading. All agents follow a shared policy. In the MARL models (Both DQN and PPO), a single policy is trained and applied to all agents, enabling them to make decisions based on their local observations while leveraging the collective learning process. The simulation (flow explained in Algorithm 2) results were evaluated based on three key metrics:

\begin{enumerate}
\item[$(i)$] Energy purchased by farms with P2P trading vs without P2P trading (grid only)
\item[$(ii)$] Energy sold by farms with P2P trading vs without P2P trading (grid only)
\item[$(iii)$] Peak hour energy demand from the grid by farms (P2P vs non-P2P)
\end{enumerate}

\subsection{PPO and DQN Learning and Stability}

\begin{figure*}
    \centering
    \begin{subfigure}{\textwidth}
        \centering
        \includegraphics[width=\textwidth]{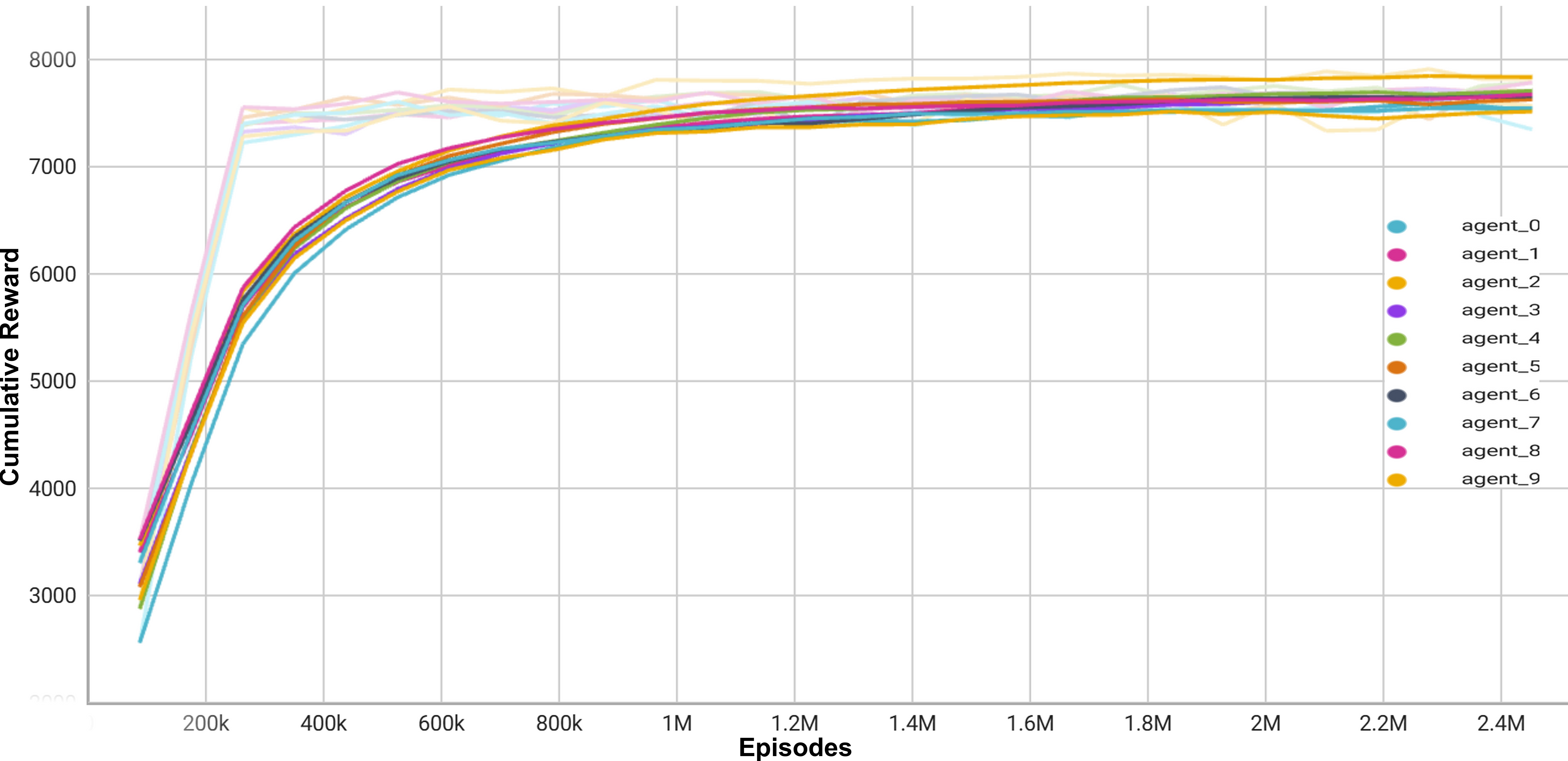}
        \caption{DQN Model Convergence}
        \label{DQNrewards}
    \end{subfigure}

    \begin{subfigure}{\textwidth}
        \centering
        \includegraphics[width=\textwidth]{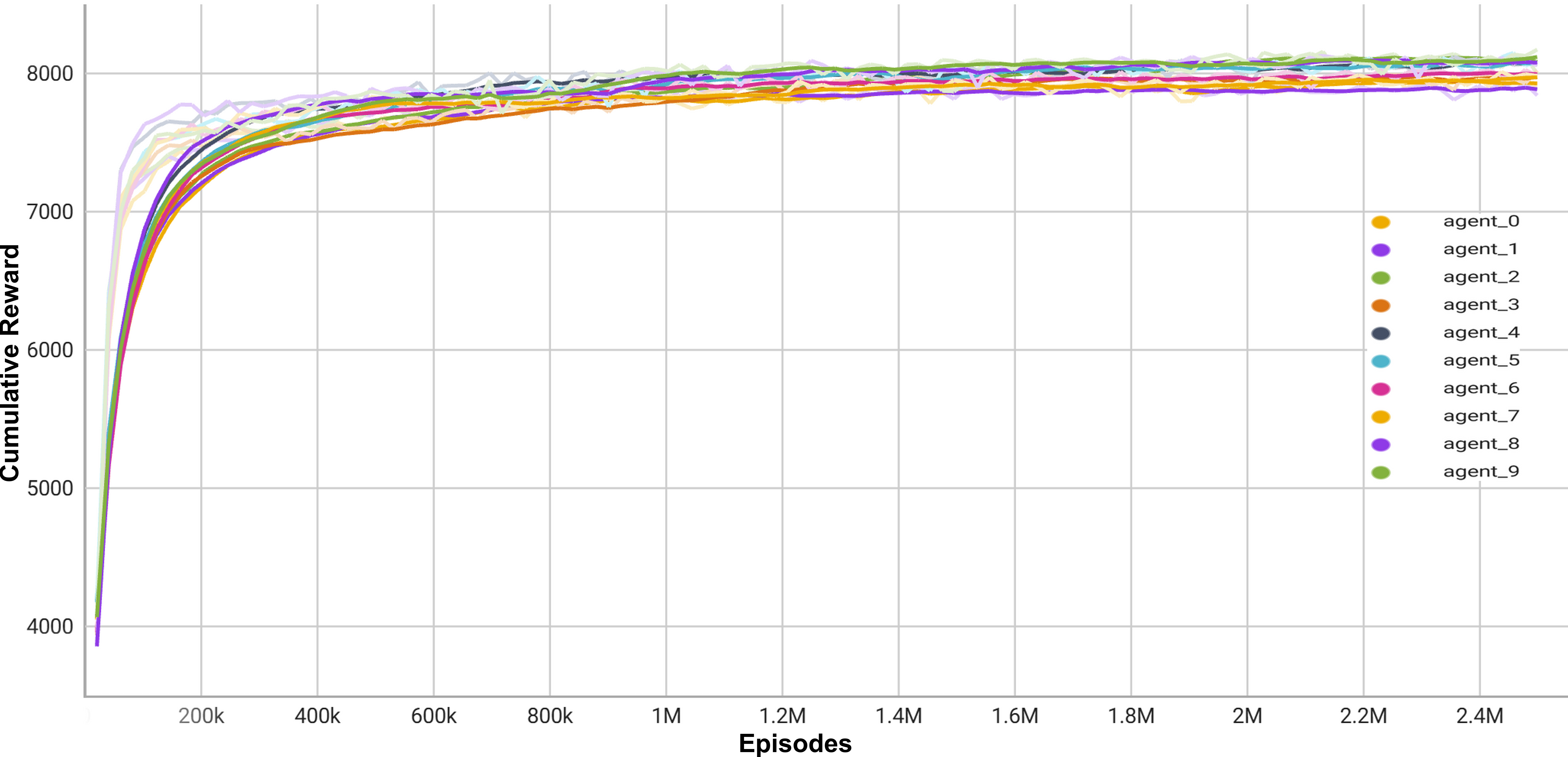}
        \caption{PPO Model Convergence}
        \label{PPOrewards}
    \end{subfigure}

    \caption{Reward Convergence for Dairy Farm Agents over 2.5M episodes}
    \label{RLConvergence}
\end{figure*}

Figures~\ref{DQNrewards} and~\ref{PPOrewards} illustrate the reward convergence over 2.5 million episodes for DQN and PPO in a multi-agent P2P energy trading environment. The training uses hourly time steps, and the curves show average performance across 10 runs per agent, with minimal variations between runs, demonstrating consistent performance. The hyperparameters used are the defaults from Stable-Baselines3 library \cite{stablebaselines3docs} \cite{rlbaselines3zoo}, detailed in \ref{hyperparameters}. Figure \ref{RLConvergence} shows each agent's learning trajectory, highlighting the optimization of cumulative rewards for energy trading strategies.

PPO demonstrates rapid initial learning, with rewards increasing sharply within the first 200,000 steps as agents explore the environment. It converges faster, stabilizing around 800,000 steps, benefiting from its policy-based approach with clipped surrogate objective and entropy-based exploration. This makes PPO well-suited for dynamic, high-dimensional environments, though its policy updates can sometimes lead to variable performance.

DQN shows a more gradual learning curve, taking about 400,000 steps to achieve similar reward levels and converging around 1 million steps. While slower to converge due to its value-based approach and epsilon-greedy exploration, DQN's off-policy nature and experience replay provide high sample efficiency. Its deterministic updates ensure smoother learning curves and more stable performance, particularly valuable in P2P energy trading where precise, discrete decision-making is crucial. These characteristics, combined with its ability to handle variable environments effectively, make DQN especially suitable for P2P energy trading applications despite its longer convergence time.

\subsection{Comparison of Proposed Models with Literature}

The results generated by PPO and DQN, based on key metrics, are presented in Table \ref{finlandresult} for results based on Finland dataset and Table \ref{irelandresult} for Ireland dataset. These tables compare the proposed PPO and DQN methods with previously published approaches, including the Rule-based simulator \cite{shah2024peer}, and Ensemble Rule-based simulator and Q-learning simulator \cite{shah2024reinforcement}. The best results (lowest for electricity purchase and peak hour demand, and highest for electricity sold) are highlighted in bold. While our study focuses on MARL-based approaches, prior work has demonstrated the relative strengths of heuristic optimization methods like Genetic Algorithms (GA) and Particle Swarm Optimization (PSO) in energy management contexts. For instance, comparative studies in smart grid phase reconfiguration show GA and PSO achieving 12-18\% cost reductions in static optimization scenarios, though with higher computational overhead than RL for dynamic environments \cite{islam2019comparison}. Similarly, hybrid PSO-RL methods in cloud load balancing highlight PSO's faster initial convergence but RL's superior adaptability to real-time fluctuations \cite{pradhan2022intelligent}. In nano-grid control, GA and PSO achieve 8-15\% efficiency gains in offline optimization but struggle with the temporal dependencies that MARL handles effectively \cite{yousaf2020comparative}. Notably, constrained portfolio trading systems combining PSO with RL \cite{almahdi2019constrained} mirror our findings: PSO excels in short-term local optimization (e.g., single-auction pricing), while RL dominates in sequential decision-making (e.g., long-term battery scheduling). This aligns with our MARL results, where DQN/PPO outperform rule-based methods by 14-55\% in dynamic metrics like peak demand reduction-a scenario where heuristic methods typically require problem-specific tuning and lack MARL's self-adaptation to unseen conditions. Future work could hybridize these approaches, using GA/PSO for initial policy warm-up or constraint handling.

\begin{table}
\caption{Comparison of Finland Results (with and without P2P Trading)}
\centering
\resizebox{\textwidth}{!}{%
\begin{tabular}{ccccc}
\hline
\textbf{Finland Data} & \textbf{Rule Based} & \textbf{RB+QL} & \textbf{DQN} & \textbf{PPO} \\
\hline
Electrcity Cost w/o P2P (Bought) (€) & 248000 & 243530 & \textbf{231950} & 241100 \\
Electrcity Cost P2P (Bought) (€) & 242000 & 237160 & \textbf{229500} & 238700 \\
Electrcity Revenue w/o P2P (Sold) (€) & 15700 & \textbf{15750} & 8870 & 7800 \\
Electrcity Revenue P2P (Sold) (€) & 16500 & 16920 & \textbf{18600} & 12400 \\
Peak Hour Demand w/o P2P (kW) & 76450 & 76200 & \textbf{52700} & 55450 \\
Peak Hour Demand P2P (kW) & 68850 & 57820 & \textbf{50250} & 53800 \\
\hline
\end{tabular}
}
\label{finlandresult}
\end{table}

\begin{table}
\caption{Comparison of Ireland Results (with and without P2P Trading)}
\centering
\resizebox{\textwidth}{!}{%
\begin{tabular}{ccccc}
\hline
\textbf{Ireland Data} & \textbf{Rule Based} & \textbf{RB+QL} & \textbf{DQN} & \textbf{PPO} \\
\hline
Electrcity Cost w/o P2P (Bought) (€) & 68030 & 66600 & \textbf{57650} & 58900 \\
Electrcity Cost P2P (Bought) (€) & 65350 & 64150 & \textbf{56100} & 57850 \\
Electrcity Revenue w/o P2P (Sold) (€) & 7520 & \textbf{7560} & 5500 & 5300 \\
Electrcity Revenue P2P (Sold) (€) & 7600 & 7750 & \textbf{8150} & 5800 \\
Peak Hour Demand w/o P2P (kW) & 12500 & 12150 & 6200 & \textbf{5200} \\
Peak Hour Demand P2P (kW) & 10900 & 10380 & 5450 & \textbf{4850} \\
\hline
\end{tabular}
}
\label{irelandresult}
\end{table}

\subsection{Models Comparison via Key Performance Indicators}

\subsubsection{Electricity Cost (Bought)}

\textbf{Finland:}  
Without P2P trading, the rule-based approach incurs the highest cost of €248,000. The RB+QL method slightly reduces this cost to €243,530. RL models show a significant improvement, with DQN reducing costs to €231,950 and PPO achieving €241,100, demonstrating better efficiency in purchasing strategies.  
With P2P trading, the costs are further reduced. The rule-based approach incurs a cost of €242,000, while RB+QL reduces this to €237,160. DQN achieves the lowest cost at €229,500, followed by PPO at €238,700.

\textbf{Ireland:}  
In Ireland, without P2P trading, the rule-based method incurs a higher electricity cost of €68,030. The RB+QL method reduces this to €66,600. RL models demonstrate significant savings, with DQN at €57,650 and PPO at €58,900.  
With P2P trading, the costs are further reduced. The rule-based approach incurs a cost of €65,350, while RB+QL reduces this to €64,150. DQN achieves the lowest cost at €56,100, followed by PPO at €57,850. These results also emphasize the cost-saving benefits of P2P trading.

\subsubsection{Electricity Revenue (Sold)}

\textbf{Finland:}  
Without P2P trading, revenue is relatively low. The rule-based model generates €15,700, with RB+QL achieving slightly higher revenue at €15,750. However, DQN and PPO generate lower revenues of €8,870 and €7,800, respectively.  
With P2P trading, revenue increases significantly. The rule-based model generates €16,500, while RB+QL achieves €16,920. DQN achieves the highest revenue at €18,600, followed by PPO at €12,400. This demonstrates the advantage of P2P trading in improving energy-selling opportunities.

\textbf{Ireland:}  
As in Finland, revenue without P2P trading is relatively low. The rule-based model generates €7,520, with RB+QL achieving slightly higher revenue at €7,560. DQN and PPO generate €5,500 and €5,300, respectively.  
With P2P trading, revenue improves. The rule-based model generates €7,600, while RB+QL achieves €7,750. DQN achieves the highest revenue at €8,150, followed by PPO at €5,800. These results highlight the revenue benefits of P2P trading.

\subsubsection{Peak Hour Demand from Grid (kW)}

\textbf{Finland:}  
Without P2P trading, the rule-based approach results in the highest peak hour demand at 76,450 kW. RB+QL slightly reduces this to 76,200 kW. DQN reduces it significantly to 52,700 kW, while PPO achieves 55,450 kW.  
With P2P trading, peak hour demand is further reduced. The rule-based approach results in 68,850 kW, while RB+QL reduces this to 57,820 kW. DQN achieves the lowest peak hour demand at 50,250 kW, followed by PPO at 53,800 kW. This demonstrates the effectiveness of P2P trading as well in reducing grid reliance during peak hours.

\textbf{Ireland:}  
In Ireland, without P2P trading, the rule-based model results in peak hour demand of 12,500 kW. RB+QL reduces this to 12,150 kW. DQN reduces this significantly to 6,200 kW, with PPO achieving a further reduction to 5,200 kW.  
With P2P trading, peak hour demand is further reduced. The rule-based approach results in 10,900 kW, while RB+QL reduces this to 10,380 kW. DQN achieves a peak hour demand of 5,450 kW, with PPO achieving the lowest at 4,850 kW. These results show the combined effectiveness of RL models and P2P trading in optimizing grid usage during peak hours.

\subsection{Improvement with P2P Trading}

To quantify the impact of peer-to-peer (P2P) trading, we calculated the percentage differences in electricity buying cost, selling revenue, and peak hour demand between scenarios with and without P2P for both the DQN and PPO approaches in Finland and Ireland. In Finland, the DQN method achieved a reduction in electricity buying cost of approximately 1.06\%, a substantial increase in selling revenue of about 109.6\%, and a decrease in peak hour demand of 4.65\%. The PPO approach in Finland showed a 1.00\% reduction in buying cost, a 59.0\% increase in selling revenue, and a 3.0\% reduction in peak demand. For Ireland, the DQN method resulted in a 2.69\% reduction in buying cost, a 48.2\% increase in selling revenue, and a 12.1\% reduction in peak hour demand. The PPO approach in Ireland demonstrated a 1.78\% reduction in buying cost, a 9.43\% increase in selling revenue, and a 6.73\% reduction in peak demand. These results clearly demonstrate that the integration of P2P trading leads to notable improvements in both economic and operational metrics. DQN consistently outperformed PPO across all key metrics in both Finland and Ireland, achieving greater reductions in buying cost and peak demand, as well as substantially higher increases in selling revenue. Thus, DQN is the best-performing model for maximizing the benefits of P2P trading.

\subsection{Ablation study: Effect of price advisor, pre-peak battery priming, and dairy-farm constraints}

\begin{table}[ht]
\caption{Ablation Study (Finland Data)}
\centering
\small
\begin{tabular}{lcccc}
\hline
\textbf{Metric} & \makecell{\textbf{Proposed}\\\textbf{Approach}} & \makecell{\textbf{No }\\\textbf{Advisor}} & \makecell{\textbf{No prepeak}\\\textbf{priming}} & \makecell{\textbf{No dairy}\\\textbf{variables}} \\
\hline
Electricity Cost w/o P2P (Bought) (€) & \textbf{231,900} & 238,600 & 235,800 & 239,900 \\
Electricity Cost with P2P (Bought) (€) & \textbf{229,400} & 236,900 & 233,900 & 237,200 \\
Electricity Revenue w/o P2P (Sold) (€) & 8,900 & \textbf{9,400} & 8,600 & 9,100 \\
Electricity Revenue with P2P (Sold) (€) & \textbf{18,700} & 14,600 & 16,200 & 15,100 \\
Peak Hour Demand w/o P2P (kW) & \textbf{52,600} & 56,800 & 54,900 & 57,400 \\
Peak Hour Demand with P2P (kW) & \textbf{50,200} & 54,700 & 52,800 & 55,300 \\
\hline
\end{tabular}
\label{tab:ablation_finland}
\end{table}

Table~\ref{tab:ablation_finland} summarizes an ablation of the price advisor, pre-peak battery priming, and dairy-specific constraints in our DQN-based MARL framework. The full model achieves the lowest costs and peaks and the highest P2P revenue. Removing the price advisor causes the largest drop in P2P revenue (-22\%) and increases costs (+€{}7{,}500) and peaks (+9.0\%). Disabling pre-peak priming moderately worsens costs (+€{}4{,}500) and peaks (+5.2\%). Removing dairy constraints further degrades performance (+€{}7{,}800 cost; -19\% revenue; +10.2\% peak), underscoring the necessity of domain-specific constraints and adaptive pricing for realistic, deployable performance.

\subsection{Summary of Analysis}

This research demonstrates the effectiveness of RL models, specifically \textbf{DQN} and \textbf{PPO}, in optimizing energy trading strategies in both P2P and non-P2P contexts. Compared to traditional rule-based and ensemble methods, RL approaches consistently outperform in terms of cost efficiency, revenue generation, and demand management, highlighting their adaptability and robustness in dynamic energy systems.

\textbf{DQN} emerges as the best-performing model for minimizing electricity costs in both P2P and non-P2P contexts for Finland and Ireland. It achieves the lowest electricity costs across all scenarios, demonstrating its ability to optimize energy purchasing strategies effectively. Additionally, DQN performs best in maximizing electricity revenue in P2P trading for both countries and excels in reducing peak hour demand in Finland for both P2P and non-P2P scenarios. \textbf{PPO}, while slightly behind DQN in cost minimization and P2P revenue generation, performs best in reducing peak hour demand in Ireland for both P2P and non-P2P scenarios.

The \textbf{RB+QL ensemble model} performs best for electricity revenue in non-P2P trading scenarios for both Finland and Ireland due to its trade-off between selling excess energy and reducing peak hour demand. However, it lacks the ability to ensure batteries are charged before high-tariff hours, limiting its effectiveness compared to RL models. In contrast, DQN and PPO use reward functions to optimize battery charging, enabling better performance in cost minimization and peak hour demand reduction.

While RL models show reduced selling revenue in non-P2P scenarios, this is offset by their superior performance in P2P trading. RL models excel in buying electricity, reducing peak hour demand, and maximizing selling revenue in P2P trading, making them the better overall approach for optimizing energy trading strategies.

\begin{figure*}
    \centering
    \begin{subfigure}{\textwidth}
        \centering
        \includegraphics[width=\textwidth]{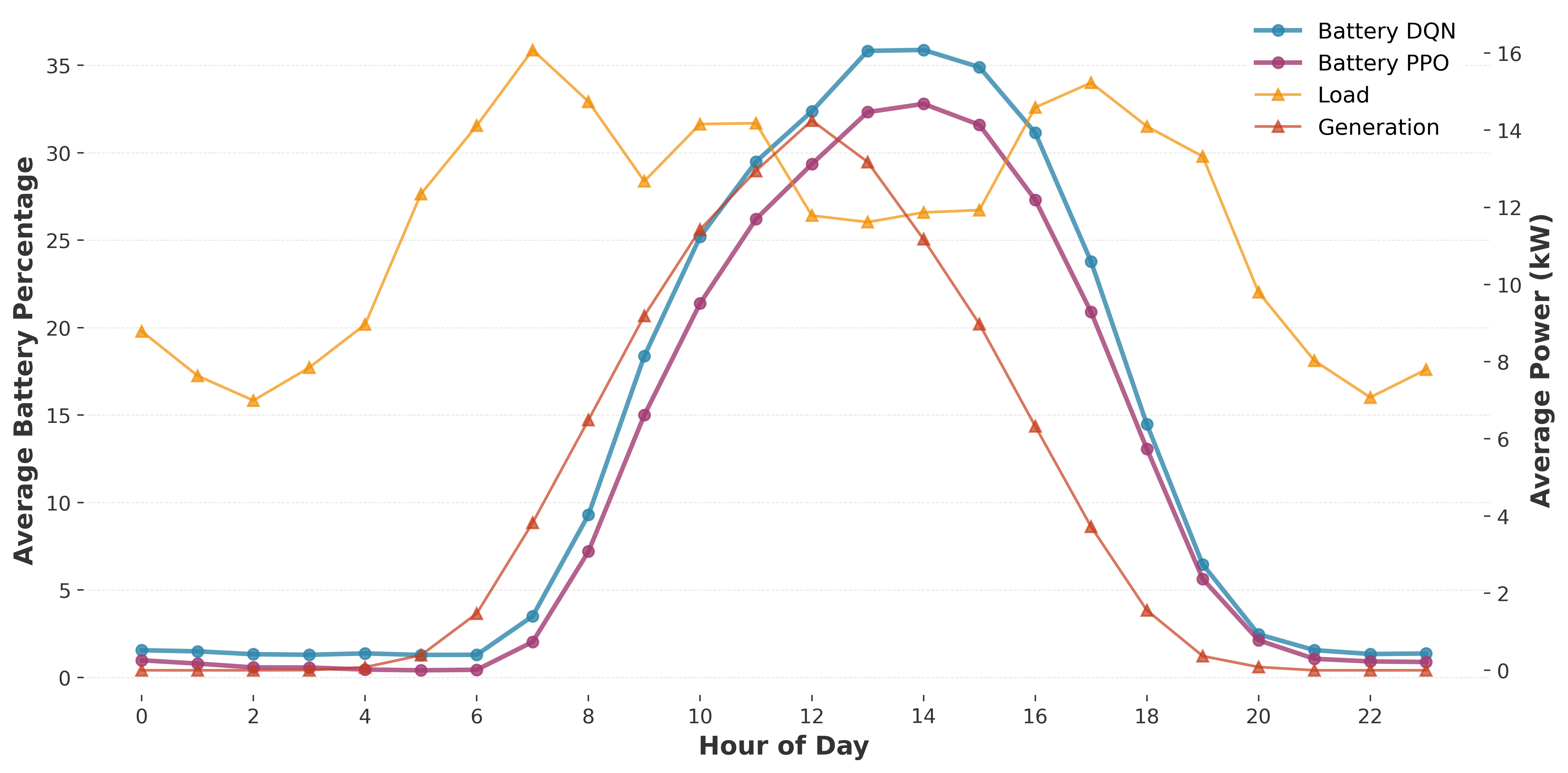}
        \caption{Finnish Farms}
        \label{FinlandProfile}
    \end{subfigure}
    \begin{subfigure}{\textwidth}
        \centering
        \includegraphics[width=\textwidth]{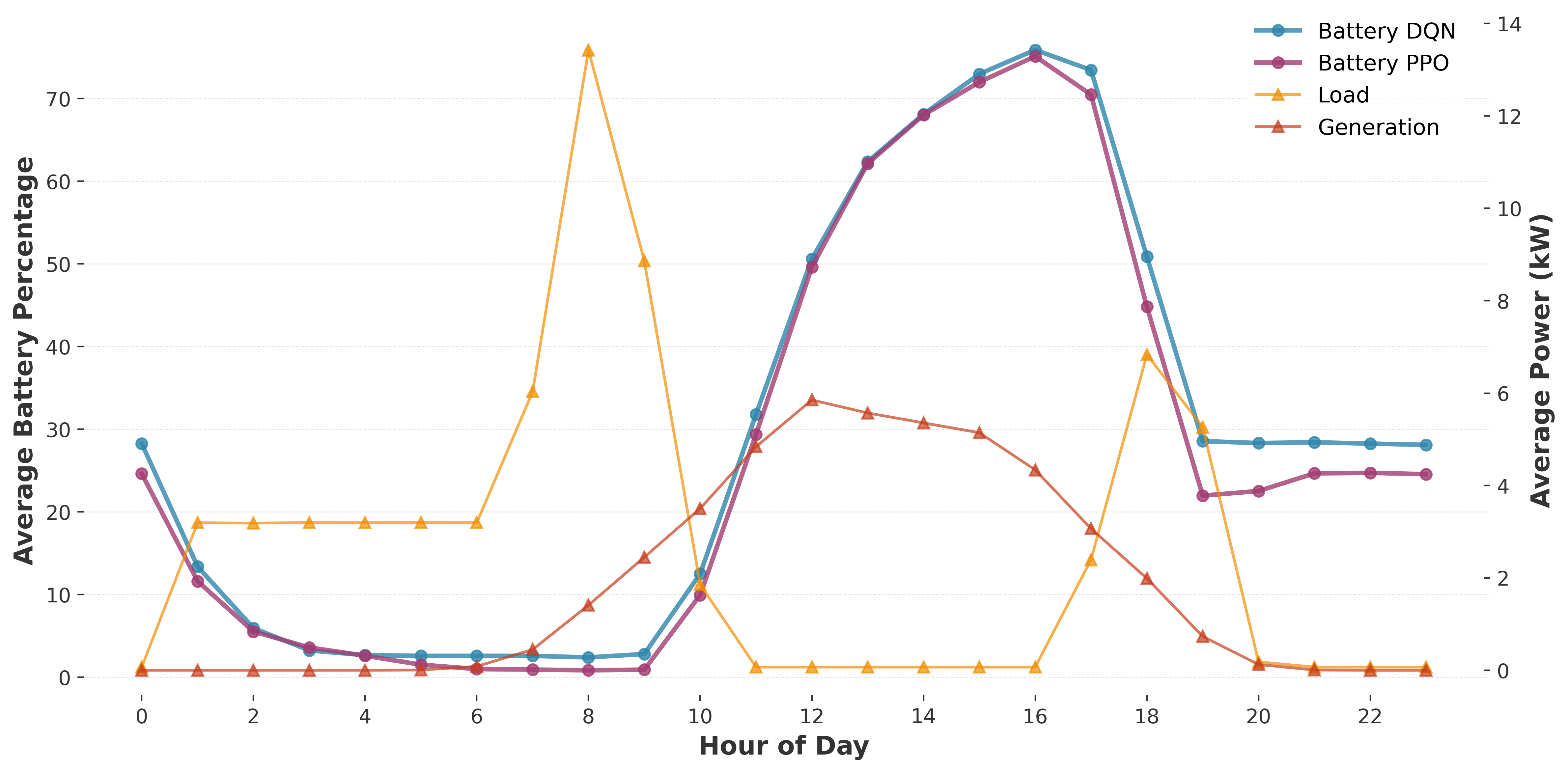}
        \caption{Irish Farms}
        \label{IrelandProfile}
    \end{subfigure}

    \caption{Typical daily patterns of battery SOC (PPO \& DQN), load, and generation, averaged across all farms over a year}
    \label{LoadnBatterypatterns}
\end{figure*}

The differences between Finland and Ireland results also stem from distinct load and generation patterns. In Finland (Figure \ref{FinlandProfile}), the load is consistent throughout the day, allowing RL models, particularly DQN, to optimize battery usage and electricity purchases predictably. In Ireland (Figure \ref{IrelandProfile}), several hours of near-zero load, as generated by the load modeling tool for Irish dairy farms, before peak demand allow batteries to fully charge using renewable generation, resulting in a more pronounced reduction in peak hour demand.

\begin{figure*}
    \centering
    \includegraphics[width=1\textwidth]{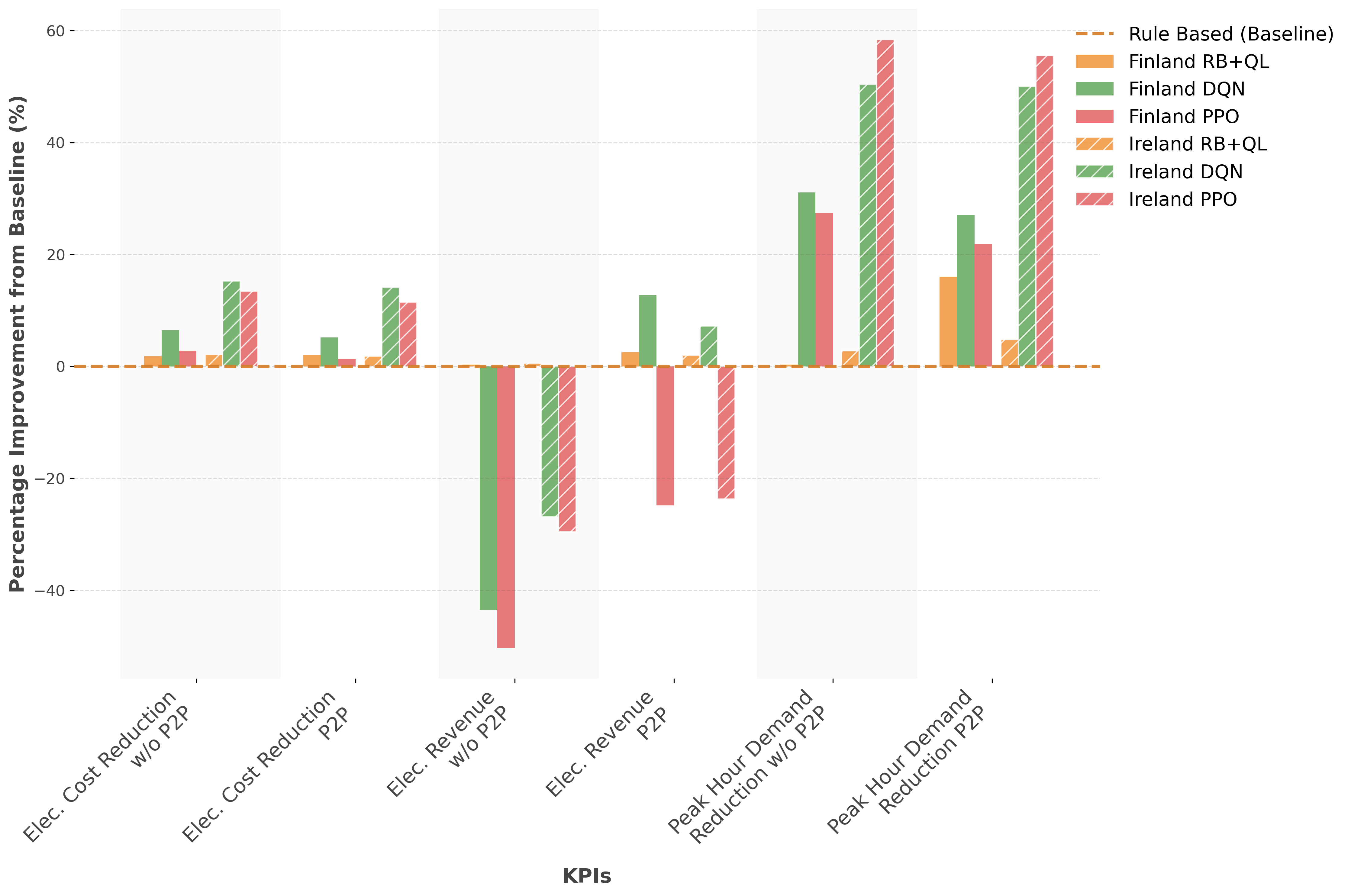}
    \caption{Comparative Percentage Results Across Models: Rule-based, RB+DQN, DQN, and PPO}
    \label{modelcompare}
\end{figure*}

Figure \ref{modelcompare} compares the models in Key Performance Indicators (KPIs). As summarized in Table \ref{finlandresult}, DQN consistently outperforms other models for Finland, achieving the lowest electricity costs, highest P2P revenue, and the greatest reduction in peak hour demand. Specifically, DQN reduces electricity costs by 6.5\% without P2P and 5.2\% with P2P compared to Rule-Based, while reducing peak hour demand by 31.1\% without P2P and 27.0\% with P2P.

For Ireland (Table \ref{irelandresult}), DQN achieves the lowest electricity costs, reducing them by 15.3\% without P2P and 14.2\% with P2P compared to Rule-Based. While PPO slightly outperforms DQN in reducing peak hour demand, DQN still achieves significant reductions of 50.4\% without P2P and 50.0\% with P2P. Overall, DQN demonstrates strong performance across KPIs, particularly in minimizing costs and peak hour demand.

In scenarios where \textbf{DQN} outperforms PPO, the performance difference is significant, particularly in cost minimization, P2P revenue generation, and peak hour demand reduction in stable environments like Finland. While PPO adapts well to volatile conditions, DQN's value-based approach consistently delivers better results. The most significant factor in lowering electricity costs, reducing peak hour demand, and increasing revenue is the implementation of \textbf{P2P energy trading}, which enhances the efficiency and effectiveness of energy management strategies, amplifying the benefits of RL models.

\subsection{Parameter Sensitivity and Scalability}

A systematic evaluation of PPO and DQN was conducted to assess parameter sensitivity and scalability. For parameter sensitivity, critical hyperparameters such as learning rate, batch size, and gamma were varied for both algorithms. PPO demonstrated stable performance across a range of learning rates, with average rewards ranging from 8058.4 to 8183.7 and standard deviations ($\sigma$) between 128.5 and 234.2. For example, with a learning rate of 0.0003, PPO achieved an average reward of 8148.3 ($\sigma$: 184.5), while increasing the learning rate to 0.001 resulted in a slightly higher average reward of 8183.7 ($\sigma$: 128.5). In contrast, DQN exhibited more pronounced sensitivity to hyperparameter changes. With a learning rate of 0.0003, DQN achieved an average reward of 4515.1 ($\sigma$: 1816.9), which increased to 6939.9 ($\sigma$: 1248.7) at a learning rate of 0.001, and further to 7214.1 ($\sigma$: 594.0) at 0.003. These results indicate that DQN’s performance is more dependent on careful hyperparameter tuning compared to PPO.

Scalability experiments assessed the performance of both algorithms as the number of agents increased. PPO maintained high average rewards and low variance even as the agent count scaled from 5 to 30. For instance, with 5 agents, PPO achieved an average reward of 8127.2 ($\sigma$: 181.7), and with 30 agents, the average reward was 8256.8 ($\sigma$: 273.2). Training time for PPO scaled linearly, from 149.1 seconds (5 agents) to 1017.1 seconds (30 agents). DQN, however, exhibited a drop in average reward and increased variance as the number of agents grew. With 5 agents, DQN’s average reward was 7951.6 ($\sigma$: 370.1), but with 30 agents, the average reward dropped to 7012.7 ($\sigma$: 2012.2). These findings suggest that PPO is more robust and scalable in multi-agent settings, while DQN’s performance degrades with increased agent count, likely due to higher communication overhead and less stable learning dynamics. The detailed tables and plots for the sensitivity and scalability analysis are provided in \ref{hyperparameters}.

\subsection{Effects of Relative Battery Size}

Analysis of the seasonal simulation results provides a comprehensive understanding of how the ratio of battery size to load and PV generation influences system performance under both P2P and non-P2P scenarios. Table \ref{batterybyseason} summarizes the average power of each component in the simulation (in kW, except for SoC) across all agents. The battery power is calculated as a quadratic average to better represent the intensity of charge and discharge cycles. The table shows that, in P2P trading, battery usage is less dependent on the load size, as evidenced by consistently higher battery power values across all seasons compared to the non-P2P case. For example, in spring, the average battery power in the P2P scenario is 2.929 kW, notably higher than the 2.046 kW observed in the non-P2P case. This trend persists throughout the year, with P2P battery power remaining above 2.8 kW in all seasons, while non-P2P values are lower and more closely aligned with the seasonal load.

Furthermore, the results indicate that the power exchanged with the main grid is strongly influenced by the ratio of battery size to PV generation. In both P2P and non-P2P cases, seasons with higher PV generation relative to load (such as summer and spring) see increased power sold to the grid and higher state of charge (SoC) values. For instance, in summer, the SoC reaches 37.007 in the non-P2P case and 35.421 in P2P, reflecting greater battery utilization and grid interaction when PV output is high. Conversely, in autumn and winter, when PV generation is much lower than the load, both the power sold to the grid and the SoC decrease, particularly in the non-P2P scenario.

The table also highlights that, in P2P trading, the battery is charged and discharged at similar rates regardless of the load, while grid transactions are more sensitive to the balance between generation and demand. This is evident from the relatively stable battery power values in P2P across all seasons, contrasted with the more variable grid power bought and sold, which closely follow the seasonal changes in PV generation and load.

\begin{table}
\centering
\caption{Seasonal simulation results: battery size impact on system performance.}
\label{batterybyseason}
\begin{tabularx}{\textwidth}{ccc*{4}{>{\centering\arraybackslash}X}*{4}{>{\centering\arraybackslash}X}}
\hline
\multirow{2}{*}{\textbf{Season}} & \multirow{2}{*}{\textbf{Load}} & \multirow{2}{*}{\textbf{PV}} & \multicolumn{4}{c}{\textbf{Non-P2P}} & \multicolumn{4}{c}{\textbf{P2P}} \\
 & & & Battery Power & Power Bought & Power Sold & SoC & Battery Power & Power Bought & Power Sold & SoC \\
\hline
Spring & 2.532 & 2.231 & 2.046 & 1.505 & 0.655 & 31.795 & 2.929 & 2.062 & 1.228 & 33.115 \\
Summer & 2.281 & 2.392 & 2.073 & 1.249 & 0.788 & 37.007 & 2.851 & 1.853 & 1.386 & 35.421 \\
Autumn & 1.883 & 0.655 & 1.696 & 1.577 & 0.143 & 19.147 & 2.928 & 2.233 & 0.364 & 35.150 \\
Winter & 2.144 & 0.800 & 1.807 & 1.747 & 0.164 & 19.442 & 2.997 & 2.337 & 0.422 & 33.629 \\
\hline
\end{tabularx}
\end{table}

These results underscore the importance of considering both battery size and the relative proportions of load and PV generation when evaluating system performance, especially in the context of P2P energy trading.

\subsection{Analysis of Battery Profiles and Load-Generation Patterns}

Figures \ref{FinlandProfile} and \ref{IrelandProfile} illustrate the average battery percentage, load, and generation profiles for Finland and Ireland, respectively. Both DQN and PPO exhibit a clear strategy of charging batteries during the near-peak period (3 PM to 5 PM) to ensure sufficient energy for peak hours (5 PM to 7 PM). This demonstrates the ability of RL models to anticipate high-demand periods and optimize battery usage effectively.

In Finland, where the load is relatively stable, DQN maintains a higher average battery percentage during the near-peak period and discharges more aggressively during peak hours, aligning with its superior performance in cost minimization and peak demand reduction. PPO also performs well but is slightly less effective in these scenarios.

In Ireland, with more variable load and generation patterns, both models adapt by charging batteries during periods of high renewable generation and discharging during peak hours. While PPO slightly outperforms DQN in reducing peak demand, DQN consistently achieves better overall performance, particularly in cost minimization and revenue generation. These results highlight the adaptability of RL models to different environmental conditions and their effectiveness in optimizing energy management strategies.

Overall, DQN is preferable for scenarios prioritizing maximum cost savings with dedicated tuning, while PPO is the superior choice for large-scale, real-world deployments demanding robustness, ease of tuning, and predictable scalability. This highlights that algorithm selection should align with the specific priorities of the energy trading application. Future work should explore hybrid or ensemble approaches to combine the strengths of both models for further improvements.

\subsection{Future Financial Impact on Prosumers and Generalization}
\label{generalization}

\begin{table}
\centering
    \caption{Future Economic Comparison of P2P vs Non-P2P (DQN, Irish Dairy Farming Community)}
\label{p2p_future_benefits}
\begin{tabular}{cp{1.7cm}p{1.7cm}p{1.7cm}p{1.7cm}p{1.7cm}p{1.7cm}p{1.7cm}}
\hline
\textbf{Year} & \textbf{Cost}\newline\textbf{Non-P2P}\newline\textbf{(€)} & \textbf{Cost}\newline\textbf{P2P}\newline\textbf{(€)} & \textbf{Saving}\newline\textbf{(€)} & \textbf{Revenue}\newline\textbf{Non-P2P}\newline\textbf{(€)} & \textbf{Revenue}\newline\textbf{P2P}\newline\textbf{(€)} & \textbf{Revenue}\newline\textbf{Diff.}\newline\textbf{(€)} & \textbf{Total}\newline\textbf{Profit}\newline\textbf{(€)} \\
\hline
2025 & 57{,}650 & 56{,}100 & 1{,}550 & 5{,}500 & 8{,}150 & 2{,}650 & 4{,}200 \\
2026 & 59{,}351 & 57{,}753 & 1{,}596 & 5{,}662 & 8{,}390 & 2{,}728 & 4{,}324 \\
2027 & 61{,}092 & 59{,}449 & 1{,}643 & 5{,}828 & 8{,}637 & 2{,}818 & 4{,}461 \\
2028 & 62{,}896 & 61{,}205 & 1{,}691 & 6{,}000 & 8{,}892 & 2{,}891 & 4{,}582 \\
2029 & 64{,}764 & 63{,}023 & 1{,}741 & 6{,}179 & 9{,}156 & 2{,}977 & 4{,}718 \\
2030 & 66{,}684 & 64{,}891 & 1{,}793 & 6{,}362 & 9{,}427 & 3{,}065 & 4{,}858 \\
2031 & 68{,}644 & 66{,}798 & 1{,}846 & 6{,}549 & 9{,}704 & 3{,}155 & 5{,}001 \\
2032 & 70{,}650 & 68{,}751 & 1{,}900 & 6{,}740 & 9{,}988 & 3{,}248 & 5{,}147 \\
2033 & 72{,}725 & 70{,}770 & 1{,}955 & 6{,}938 & 10{,}281 & 3{,}344 & 5{,}299 \\
2034 & 74{,}893 & 72{,}880 & 2{,}014 & 7{,}145 & 10{,}588 & 3{,}443 & 5{,}456 \\
\hline
\textbf{Total} & -- & -- & \textbf{€17{,}727} & -- & -- & \textbf{€30{,}306} & \textbf{€48{,}032} \\
\hline
\end{tabular}
\end{table}

In recent years, governments across Europe have introduced a wide range of initiatives to accelerate the transition to renewable energy at the household level. Support mechanisms such as grants, tax rebates, and feed--in tariffs are designed to reduce the upfront burden of investments in green energy, specifically in rooftop solar PV systems. In Ireland, policy proposals such as those outlined by Malik et al.~\cite{malik2025towards} highlight the importance of empowering prosumers through P2P frameworks and the role of tailored incentives in enabling decentralised participation. In addition to grants such as the Targeted Agricultural Modernisation Scheme (TAMS), government levies and grid tariff structures significantly affect the economics of prosumer adoption. Recent techno-economic studies indicate that with current grant levels, the payback period (PBP) for a typical PV-battery system can average around five to six years \cite{bakht2025techno}, depending on the scale of self-consumption and surplus exported to the grid. Since this analysis is based solely on grid feed-in revenues, the comparative results in The following results suggest that enabling prosumers to participate in P2P trading would lower effective payback times further, as higher revenues and system-wide efficiencies compound the benefits over both five- and ten-year horizons.

The comparison of future economic benefits associated with P2P participation versus conventional grid interaction over five and ten years is presented in Table \ref{p2p_future_benefits}. It is important to note that these values represent the aggregated results for a full community of ten prosumer dairy farms rather than a single individual farm; unit-level estimates can be derived by scaling the results accordingly. Using the DQN policy, P2P trading materially improves prosumer economics relative to a non-P2P baseline. Calibrated to Ireland data, the DQN schedule reduces annual purchase costs from~\euro57{,}650 to~\euro56{,}100 and raises annual sales revenues from~\euro5{,}500 to~\euro8{,}150, implying a 2025 net benefit of~\euro4{,}200 (cost saving~\euro1{,}550 plus revenue uplift~\euro2{,}650).

To reflect evolving market conditions, these deltas are scaled each year based on the work of Faiud et al.~\cite{faiud2024evaluating}, which projects electricity prices rising from 38.69\,c/kWh in 2025 to 50.26\,c/kWh in 2034. This corresponds to an annualised increase of roughly 2.8\% per year, computed directly from the electricity price growth path. Incorporating this escalation yields cumulative nominal gains over five years of~\euro22{,}274 (cost savings~\euro8{,}220; revenue~\euro14{,}054) and over ten years of~\euro48{,}032 (cost savings~\euro17{,}726; revenue~\euro30{,}306). Expressed in constant euros using ECB inflation projections, the 10-year net benefit is~\euro43{,}094 under the staff baseline (2025: 2.0\%, 2026: 1.6\%, 2027: 2.0\%, then 2\%) and~\euro42{,}937 under the ECB Survey of Professional Forecasters path (2.1\%, 1.9\%, 2.0\%, then 2\%). Rising electricity prices amplify the advantage of P2P: relative to a counterfactual with prices fixed at 2025 levels, the 10-year nominal net benefit is higher by about~\euro6{,}032, with roughly two-fifths coming from larger cost savings and three-fifths from higher sales revenues.

Overall, these results indicate that P2P trading not only shifts the operating point toward lower net expenditure in year one, but also compounds into substantial multi-year value as retail tariffs escalate, with benefits robust across plausible inflation scenarios.\\

Evidence of generalization based on the same model from a heterogeneous community (dairy + residential) with integrated forecasting shows the proposed policy-market interface retains performance under agent diversity and uncertainty in \cite{shah2025uncertainty}. The MARL framework with a price advisor and double-auction market mechanism remained robust under agent heterogeneity and forecast uncertainty, consistently improving the same KPIs, lowering energy purchase costs, reducing peak demand, and increasing sales revenues. Relative to grid-only baselines, the model reduces energy purchase costs by up to 5.7\% without P2P trading and 3.2\% with P2P trading, while increasing electricity sales revenue by 6.4\% and 44.7\%, respectively. Additionally, peak-hour grid demand is reduced by 38.8\% without P2P and 45.6\% with P2P. This supports the method’s applicability beyond the specific case studies.\\
Moreover, the model, trained on an Irish dataset and validated with a Finnish dataset, exhibited strong out-of-distribution performance. This cross-country transferability, achieved with minimal architectural or hyperparameter adjustments, suggests the approach captures generalizable control and market-response patterns, rather than overfitting to specific regional conditions.

\section{Future Work and Challenges}
In addition to exploring the ensembling of DQN and PPO, several future research directions and challenges must be addressed to enhance the implementation of P2P energy trading using MARL in dairy farms:

\textit{Translating Simulations to Real-World Scenarios}\\
Simulations provide a controlled environment, but real-world deployment faces challenges like weather variations, equipment failures, and human intervention, which are difficult to model accurately.

\textit{Dairy Farm-Specific Research}\\
Current P2P energy trading and MARL approaches are designed for general use cases and do not account for the unique energy consumption patterns, equipment needs, and renewable energy potential of dairy farms. Synchronized load patterns among dairy farms can significantly reduce the opportunities for peer-to-peer energy trading, as simultaneous peaks in demand limit the availability of surplus energy for exchange within the community. To address this, future work could explore the use of asynchronous control strategies, where flexible loads are scheduled at different times across farms, or virtual aggregation methods that coordinate multiple participants as a single trading entity. These approaches have the potential to diversify supply and demand profiles, enhance local trading opportunities, and improve overall system efficiency.

\textit{Adoption of Renewable Energy and P2P Trading}\\
Many dairy farms lack the infrastructure, expertise, or financial resources to adopt renewable energy and P2P trading systems. Cultural and behavioral shifts are also needed to build trust in these systems.

\textit{Line Losses in Rural Settings}\\
Rural dairy farms are often spread out, leading to significant line losses that reduce system efficiency. Future research should incorporate line losses into MARL models to better reflect rural energy trading realities.

\textit{Battery Management and Optimization}\\
Investigating battery cycles, degradation, and sizing is critical for optimizing battery usage, ensuring system longevity, and improving cost-effectiveness.

\textit{Expansion to Rural Communities}\\
The proposed system could be extended to other rural communities with similar energy needs, requiring adaptation to diverse energy consumption patterns and renewable energy potential.

\textit{Load and Generation Prediction}\\
Integrating predictive features, such as machine learning models for time-series forecasting, could improve decision-making by forecasting load, generation, and battery usage in advance. Accurate forecasting is particularly important in heterogeneous communities, where different types of prosumers (e.g., dairy farms and residential houses) exhibit diverse consumption and generation patterns. Enhancing the forecasting component of our framework will be crucial for supporting robust and adaptive P2P trading in such settings.

\textit{Heterogeneity and Generalization}\\
While our initial work focused on a relatively homogeneous group of dairy farms, real-world energy communities are often composed of diverse prosumers with varying operational profiles and energy needs. Extending our MARL-based framework to handle this heterogeneity is an important direction for future research. This will involve developing and testing individualized or parameterized policies that can generalize across different types of agents, ultimately improving the scalability and applicability of our approach to broader rural and mixed communities.

\textit{Policy and Incentive Design}\\
Regulatory and policy challenges may hinder the adoption of decentralized trading models. Future research should focus on designing supportive policies and incentives, collaborating with policymakers, energy providers, and farmer organizations.

\textit{Data Availability and Quality}\
High-quality data is essential for MARL models, but many dairy farms lack the necessary sensors and infrastructure, leading to challenges with data sparsity and noise. To address these limitations, we suggest several strategies for future work, including the use of data augmentation techniques, transfer learning from similar domains, and the integration of synthetic data generated from realistic simulations to supplement real-world measurements. Additionally, robust training methods such as noise-injection during learning and regularization can help improve model resilience to imperfect data. Ensuring data privacy and security is also a critical challenge; therefore, in accordance with the rules of the distributed P2P energy trading system, only minimal data (load, generation, and price) is shared with the auctioneer. Future work should explore anonymization techniques or further minimal information-sharing strategies to enhance privacy.

Addressing these challenges and exploring these directions will be key to scaling and deploying MARL-based P2P energy trading systems in dairy farms and other rural communities.

\section{Conclusion}

This research highlights the effectiveness of integrating Multi-Agent Reinforcement Learning (MARL), specifically Proximal Policy Optimization (PPO) and Deep Q-Networks (DQN), into distributed peer-to-peer (P2P) energy trading systems for dairy farming communities. By leveraging auction-based market clearing mechanisms and a price advisor agent integrated with load and battery management, the proposed approach significantly outperforms traditional methods. The key contributions and findings are summarized below:

\begin{itemize}
    \item P2P energy trading leads to overall improvements in reducing electricity costs, increasing electricity selling revenue, and lowering peak hour electricity demand.
    \item The integration of MARL in P2P energy trading outperforms traditional rule-based methods across all key metrics.
    \item \textbf{Electricity Costs:} DQN achieves the lowest electricity costs in Ireland, reducing costs by 14.2\% with P2P trading compared to the Rule-Based method and outperforming PPO by 2.7\%. In Finland, DQN reduces costs by 5.16\% and outperforms PPO by 3.8\%.
    \item \textbf{Peak Hour Demand:} PPO achieves the lowest peak hour demand in Ireland, reducing it by 55.5\% with P2P trading compared to the Rule-Based method, while DQN achieves a 50.0\% reduction. In Finland, DQN reduces peak hour demand by 27.02\%, outperforming PPO, which achieves a 21.86\% reduction.
    \item \textbf{Electricity Revenue:} DQN outperforms all other models in both Ireland and Finland under P2P trading, increasing revenue by 7.24\% in Ireland and 12.73\% in Finland compared to the Rule-Based method.
    \item \textbf{Model Suitability:} DQN is more stable and consistent than PPO for P2P energy trading. Both models significantly outperform traditional methods, showcasing the potential of reinforcement learning in transforming energy trading systems.
    \item \textbf{Future Potential:} The complementary strengths of DQN and PPO suggest that hybrid approaches or context-specific model selection could further enhance performance in future applications.
\end{itemize}

This research establishes a foundation for applying RL in P2P energy trading within the agriculture sector. Although demonstrated for rural dairy farms, the framework is applicable to similar rural or agricultural contexts. P2P trading benefits may be limited when participants have synchronized energy use or existing storage solutions, but this study draws attention to the often overlooked high electricity demand in agriculture, highlighting the need for targeted research and policy. Future work will extend this approach to other sectors and explore hybrid solutions for improved rural energy resilience. By tailoring RL strategies to the unique environment, this work enables more efficient, adaptable, and sustainable energy trading in dairy farming and beyond.

\section*{Acknowledgment}
This publication has emanated from research conducted with the financial support of Research Ireland under Grant number [21/FFP-A/9040].

\bibliography{bibliography}

\newpage

\appendix
\section{}
\label{loadandbatterymanagement}

\textbf{Electricity Price and Battery Management}\\

Equation \ref{lambdabuy} defines the electricity price for electricity purchased from the grid, represented by $\lambda_{\text{buy}}$, which is categorized into three levels: night, day, and peak, based on the location of the farms.

\begin{equation}
\label{lambdabuy}
\lambda_{\text{buy}} = \text{night, day, peak}
\end{equation}

\textbf{Energy Availability and Battery Management}\\

To charge or discharge the batteries and control the usage of RE resources, the following equations have been formulated. Equation \ref{usbalecapacity} computes the present usable battery capacity ($B_{\text{uc}}$) using current battery capacity, $B_{\text{cc}}$ and the maximum battery discharge capacity ($\max(B_{\text{dcap}})$). If $B_{\text{cc}}$ is less than ($\max(B_{\text{dcap}})$), $B_{\text{uc}}$ equals $B_{\text{cc}}$, otherwise, $\max(B_{\text{dcap}})$.

\begin{equation}
\label{usbalecapacity}
B_{\text{uc}} = 
\begin{cases} 
B_{\text{cc}}, & \text{if } B_{\text{cc}} < \max(B_{\text{dcap}}) \\
\max(B_{\text{dcap}}), & \text{otherwise}
\end{cases}
\end{equation}

Equation \ref{totalgeneration} evaluates the total energy generation of the farm ($E_{\text{tot}}$) by summing up the energy generated by the PV system ($E_{\text{pv}}$), wind turbine ($E_{w}$), and $B_{\text{uc}}$. If the PV system, wind turbine, or battery does not generate any energy, $E_{\text{tot}}$ is zero. $E_{\text{tot}}$, $E_{\text{pv}}$, $E_{w}$, and $B_{\text{uc}}$ cannot be less than zero.

\begin{equation}
\label{totalgeneration}
E_{\text{tot}} = E_{\text{pv}} + E_{w} + B_{\text{uc}}
\end{equation}

Equation \ref{equ4} determines the battery operation when RE resources and batteries are available on the farm. If total energy generation ($E_{\text{tot}}$) is higher than the farm load ($E_{l}$), and the current battery percentage ($\text{SoC}$) is less than 90\%, the battery is charged or discharged based on the difference between $E_{\text{tot}}$ and $E_{l}$, and the charging capacity ($B_{\text{ccap}}$). Limiting the battery charge to 90\% helps to prevent battery overheating and prolong the lifespan of the batteries \cite{saxena2016cycle}. If the available excess energy is less than what is required to charge the battery, only the battery is charged with the available quantity; otherwise, the surplus energy is also sold in the market.

\begin{equation}
\label{equ4}
\begin{aligned}
& \text{if } E_{\text{tot}} > E_{l} \text{ \& } \text{SoC} < 90: \\
& \quad 
\begin{cases}
\begin{aligned}
& E_{\text{tot}} - E_{l} < B_{\text{cp}}, \text{ charge}=1 \\
& E_{\text{tot}} - E_{l} > B_{\text{cp}}, \text{ charge } = 1 \& \text{ sell } = 1
\end{aligned}
\end{cases}
\end{aligned}
\end{equation}

\begin{equation}
\label{equ5}
\begin{aligned}
& \text{if } E_{\text{tot}} < E_{l}: \\
& \quad 
\begin{cases}
\text{SoC} > 20 \&  \lambda_{\text{buy}} = \text{night}, \text{ Buy}=1 \& \text{ charge}=0 \\
(\text{SoC} < 50 \& \lambda_{\text{buy}} = \text{night}) \text{ or} \\
(\text{SoC} < 20 \& \lambda_{\text{buy}} < \text{peak}), \text{ Buy}=1 \& \text{ charge}=1
\end{cases}
\end{aligned}
\end{equation}

Equation \ref{equ5} determines whether to charge/discharge the battery or purchase energy from the grid based on renewable energy generation ($E_{\text{tot}}$), battery percentage ($\text{SoC}$), and energy price ($\lambda_{\text{buy}}$). Night, day, and peak refer to off-peak, normal, and peak electricity tariffs. If $E_{\text{tot}}$ is less than the farm load, the battery level is above 20\% (minimum allowed by the simulator to enhance battery life \cite{saxena2016cycle}), and the energy price is not low, energy is purchased from the community without charging the battery. If the battery level is below 50\% and the energy price is low, or below 20\% and the price is not peak, the battery is charged as it is cost-effective, and energy is purchased from the grid.

If no battery is present, Equation \ref{equ6} decides whether to buy or sell energy based on RE generation ($E_{\text{tot}}$) and farm load. Energy is sold if $E_{\text{tot}}$ exceeds the farm load and bought if it is less.

\begin{equation}
\label{equ6}
\text{if } RE = 1 \text{ \& } \text{bat} = 0: \quad
\begin{cases}
\text{if } E_{\text{tot}} > E_{l}, \text{ sell}=1 \\
\text{if } E_{\text{tot}} < E_{l}, \text{ buy}=1
\end{cases}
\end{equation}

\begin{equation}
\label{equ7}
\begin{aligned}
& \text{if } RE = 0 \text{ \& } \text{bat} = 1: \\
& \quad 
\begin{cases}
\begin{aligned}
& \text{if } \text{SoC} > 20 \text{ \& } \lambda_{\text{buy}} = \text{night}, && \text{ Buy}=1 \& \text{ charge}=0 \\
& \text{if } (\text{SoC} < 20 \text{ \& } \lambda_{\text{buy}} < \text{peak}), && \text{ Buy}=1 \& \text{ charge}=1
\end{aligned}
\end{cases}
\end{aligned}
\end{equation}

Equation \ref{equ7} determines whether energy should be purchased externally and whether the battery should be charged or not depending on the current battery level ($\text{SoC}$) and energy price purchased from the grid ($\lambda_{\text{buy}}$). If $\text{SoC}$ is greater than 20\% and the energy price is not off-peak, energy is purchased externally, and the battery is not charged, otherwise, the battery is charged.

\begin{equation}
\label{batterypercent}
\text{SoC} =
\begin{cases}
\text{SoC} + \frac{(B_{\text{ccap}} \times \text{time})}{B_{c}} \times 100, & \text{charge}=1\text{, where time = 1 hr} \\
\text{SoC} - B_{\text{dp}}, & \text{discharge}=1\text{, where time = 1 hr}
\end{cases}
\end{equation}

The battery percentage ($\text{SoC}$) is updated based on the battery's charging or discharging status as in Equation \ref{batterypercent}. During charging, $\text{SoC}$ is increased by the charging capacity ($B_{\text{ccap}}$) divided by the total battery capacity ($B_{c}$). During discharging, $\text{SoC}$ is decreased by the battery discharge percentage ($B_{\text{dp}}$). The battery charging capacity and discharging percentage are calculated using Equations \ref{chargingpower} and \ref{chargingpercentage}.

When RE is generated, the battery charging capacity ($B_{\text{ccap}}$) is adjusted depending on the energy surplus. If the difference between the total energy generated ($E_{\text{tot}}$) and the farm load is less than the maximum charging capacity, then $B_{\text{ccap}}$ is adjusted to match the energy surplus. $\text{SoC}$ is updated accordingly.

\begin{equation}
\label{chargingpower}
B_{\text{ccap}} = 
\begin{cases}
\max(B_{\text{ccap}}), & \text{if } (RE = 1 \text{ and } E_{\text{tot}} - E_{l} > \max(B_{\text{ccap}})) \text{ or } RE = 0 \\
E_{\text{tot}} - E_{l}, & \text{if } RE = 1 \text{ and } E_{\text{tot}} - E_{l} < \max(B_{\text{ccap}})
\end{cases}
\end{equation}

\begin{equation}
\label{chargingpercentage}
B_{dp} = \frac{B_{uc}}{B_c}\times 100
\end{equation}

Finally, Equation \ref{excessenergy} calculates the excess energy ($E_{s}$) when energy generation exceeds the farm load. If the battery level ($\text{SoC}$) is greater than or equal to 90\%, the excess energy is calculated by subtracting the farm load from the total energy generated. If the battery level is less than 90\%, the excess energy is reduced based on the charging capacity.

\begin{equation}
\label{excessenergy}
E_{s} = 
\begin{cases}
E_{\text{tot}} - E_{l}, & \text{if SoC} \geq 90 \\
(E_{\text{tot}} - E_{l}) - B_{\text{ccap}}, & \text{if } (E_{\text{tot}} - E_{l}) > B_{\text{ccap}} \text{ and SoC} < 90
\end{cases}
\end{equation}

Finally, Equation \ref{energybuy} calculates the energy to be purchased. The energy price is determined based on the energy surplus and the battery state. If energy generation is less than the farm load, energy must be purchased.

\begin{equation}
\label{energybuy}
E_b = 
\begin{cases}
E_l, & \text{if } RE = 0 \text{ and } \text{bat} = 0 \\
E_l - (E_{\text{tot}} + B_{\text{uc}}), & \text{if } E_{\text{tot}} < E_l \text{ and } \lambda_{\text{buy}} > \text{night} \text{ and SoC} > 20 \\
(E_l - E_{\text{tot}}) + B_{\text{ccap}}, & \text{if } \lambda_{\text{buy}} < \text{day} \text{ and SoC} \leq 20 \\
E_l - E_{\text{tot}}, & \text{if } RE = 1 \text{ and } \text{bat} = 0 \\
E_l - B_{\text{uc}}, & \text{if } RE = 0 \text{ and } \text{bat} = 1 \text{ and SoC} > 20 \text{ and } \lambda_{\text{buy}} > \text{night} \\
E_l + B_{\text{ccap}}, & \text{if } RE = 0 \text{ and } \text{bat} = 1 \text{ and SoC} \leq 20 \text{ and } \lambda_{\text{buy}} < \text{night}
\end{cases}
\end{equation}

\section{}

\begin{algorithm}
\caption{Market Clearing Algorithm}
\label{auction}
\begin{algorithmic}
\State \textbf{Step 1.} Allocate buy and sell order books $O_b$ and $O_s$ at auction period $t$
\State \textbf{Step 2.} Initialize $i = j = 0$
\State \textbf{Step 3.} \While {True}
\State \textbf{Step 4.} Match the trading energy: $Q_t = \min(Q_{\beta,bi}, Q_{\sigma,si})$
\State \textbf{Step 5.} Calculate the trading price: $P_t = (P_{\beta,bi} + P_{\sigma,si})/2$
\State \textbf{Step 6.} Update buy order book: $O_{bi} \leftarrow O_{bi} - Q_{\beta,bi}$
\State \textbf{Step 7.} \If{$O_{bi} = 0$}
\State \textbf{Step 8.} $i \leftarrow i + 1$
\EndIf
\State \textbf{Step 9.} Update sell order book: $O_{sj} \leftarrow O_{sj} - Q_{\beta,sj}$
\State \textbf{Step 10.} \If{$O_{sj} = 0$}
\State \textbf{Step 11.} $j \leftarrow j + 1$
\EndIf
\State \textbf{Step 12.} \textbf{break if} $O_s$ or $O_b$ = 0
\EndWhile
\State \textbf{Step 13.} Balance unmatched quantity at ToU $\lambda_{buy}$ and FiT $\lambda_{sell}$.
\end{algorithmic}
\end{algorithm}

\section{}
\label{agentsetup}

The RL algorithms (PPO and DQN) are initialized with the following components:\\

\textbf{Agent Setup}
The environment supports multiple agents, each representing a farm. The \texttt{number\_agents} parameter defines the number of agents, each with a unique identifier. The \texttt{agent\_name\_mapping} dictionary maps agent names to indices for easy access to agent-specific data like energy load and generation profiles.\\

\textbf{Energy Profiles}
The environment has been trained on two separate electricity load profiles and generation profiles using Ireland and Finland datasets, representing the hourly energy consumption and generation of each agent, respectively. These profiles span a full year (can be increased or decreased in testing) and update every hour, reflecting dynamic energy usage and generation, influencing agent decisions.\\

\textbf{Battery and Grid Interaction}
Each agent has a battery, modeled as a percentage with an initial value of 100\%. The charging and discharging capacity and total battery capacity define the limits for energy storage and discharge. Grid prices for buying and selling energy are based on the time of day, with higher tariffs during peak hours as seen in Equation \ref{lambdabuy} (e.g., 17-19 hours).\\

\textbf{Action Space}
Each agent can choose one of eight discrete actions per time step:
\begin{itemize}
    \item \textbf{Charge \& Buy}: Charge the battery and buy energy from the grid, if required based on the rules of reward function.
    \item \textbf{Buy}: Buy energy from the grid when generation is insufficient.
    \item \textbf{Sell}: Sell excess energy to the grid.
    \item \textbf{Discharge \& Sell}: Discharge battery to sell excess energy.
    \item \textbf{Discharge \& Buy}: Discharge battery and buy energy from the grid, specifically in peak hours (if needed).
    \item \textbf{Self}: Use only self-generated energy to meet demand.
    \item \textbf{Self \& Charge}: Use self-generated energy and charge the battery from seller.
    \item \textbf{Self \& Discharge}: Use self-generated energy and discharge the battery to sell.
\end{itemize}

\textbf{State Representation}
Each agent's state (observation spaces) is represented as a four-dimensional vector:
\begin{itemize}
    \item \textbf{Load}: The agent's (farm) energy demand at the current time step.
    \item \textbf{Generation}: The agent's (farm) renewable generation at the current time step.
    \item \textbf{Battery SoC}: The battery level as a percentage of its total capacity.
    \item \textbf{Time of day}: The current hour, affecting grid pricing.
\end{itemize}

\section{}
\label{statestransition}

\textbf{State Transition Function for Agent-Environment Interactions}

\textbf{State vector for each agent at time $t$:}

\[
s_t = [L_t, G_t, \mathrm{SoC}_t, h_t, \mathrm{ISP}_t, \mathrm{IBP}_t]
\]

where:
\begin{itemize}
    \item $L_t$ = load (kWh)
    \item $G_t$ = generation (kWh)
    \item $\mathrm{SoC}_t$ = battery state-of-charge (\%)
    \item $h_t$ = hour of day ($0$--$23$)
    \item $\mathrm{ISP}_t$, $\mathrm{IBP}_t$ = internal selling/buying prices
\end{itemize}

\textbf{Constants:}
\begin{align*}
    C_{\text{BAT}} &= 13.5 \quad \text{(battery capacity, kWh)} \\
    C_{\text{CHARGE}} &= 5 \quad \text{(max charge/discharge per step, kWh)} \\
    \eta &= \frac{100}{C_{\text{BAT}}} \quad \text{(kWh to \% conversion)} \\
    \lambda_{\text{PEAK}} &= 0.66 \\
    \lambda_{\text{DAY}} &= 0.44 \\
    \lambda_{\text{NIGHT}} &= 0.22 \\
    \lambda_{\text{FIT}} &= 0.135
\end{align*}

\textbf{Actions:}
\begin{align*}
    0 &: \text{CHARGE\_AND\_BUY} \\
    1 &: \text{BUY} \\
    2 &: \text{SELL} \\
    3 &: \text{DISCHARGE\_AND\_SELL} \\
    4 &: \text{DISCHARGE\_AND\_BUY} \\
    5 &: \text{SELF} \\
    6 &: \text{SELF\_AND\_CHARGE} \\
    7 &: \text{SELF\_AND\_DISCHARGE}
\end{align*}

\begin{algorithmic}[1]
\State \textbf{Input:} $s_t = [L, G, \mathrm{SoC}, h, \mathrm{ISP}, \mathrm{IBP}]$, action $a_t$
\State \textbf{Constants:} $C_{\text{BAT}}$, $C_{\text{CHARGE}}$, $\eta = 100/C_{\text{BAT}}$
\If{$17 \leq h < 19$} 
    \State $\tau \gets$ P, $\lambda_{\text{buy}} \gets 0.66$
\ElsIf{$15 \leq h < 17$ or $23 \leq h$ or $h < 8$} 
    \State $\tau \gets$ N/NP, $\lambda_{\text{buy}} \gets 0.22$
\Else 
    \State $\tau \gets$ D, $\lambda_{\text{buy}} \gets 0.44$
\EndIf
\State $\text{buy\_kWh} \gets 0$, $\text{sell\_kWh} \gets 0$, $\mathrm{SoC}_{\text{new}} \gets \mathrm{SoC}$
\If{$a_t = 0$} 
    \If{$(\mathrm{SoC} \leq 50$ and $G < L)$ or $(\mathrm{SoC} \leq 50$ and $\tau \in \{\text{N}, \text{NP}\})$}
        \State $\Delta\% \gets \min(C_{\text{CHARGE}}, \max(G-L,0)) \cdot \eta$
        \State $\mathrm{SoC}_{\text{new}} \gets \min(\mathrm{SoC} + \Delta\%, 100)$
        \State $\text{buy\_kWh} \gets \max(L-G, 0) + C_{\text{CHARGE}}$
    \Else
        \State $\text{buy\_kWh} \gets \max(L-G, 0)$
    \EndIf
\ElsIf{$a_t = 1$}
    \If{$G < L$ and $\mathrm{SoC} < 10$ and $\tau \neq \text{N}$}
        \State $\text{buy\_kWh} \gets L-G$
    \EndIf
\ElsIf{$a_t = 2$}
    \If{$G > L$}
        \If{$\mathrm{SoC} \geq 90$ or $(\mathrm{SoC} \geq 20$ and $\tau = \text{P})$}
            \State $\text{sell\_kWh} \gets G-L$
        \EndIf
    \EndIf
\ElsIf{$a_t = 3$}
    \If{$G \geq L$ and $((\mathrm{SoC} \geq 20$ and $\tau = \text{P})$ or $(\mathrm{SoC} \geq 90$ and $\tau \neq \text{P}))$}
        \State $kWh_{\text{bat}} \gets \min(C_{\text{CHARGE}}, \mathrm{SoC} \cdot C_{\text{BAT}} / 100)$
        \State $\mathrm{SoC}_{\text{new}} \gets \mathrm{SoC} - kWh_{\text{bat}} \cdot \eta$
        \State $\text{sell\_kWh} \gets (G-L) + kWh_{\text{bat}}$
    \EndIf
\ElsIf{$a_t = 4$}
    \If{$G < L$ and $\mathrm{SoC} \geq 10$}
        \State $\text{deficit} \gets L-G$
        \State $kWh_{\text{bat}} \gets \min(C_{\text{CHARGE}}, \text{deficit}, \mathrm{SoC} \cdot C_{\text{BAT}} / 100)$
        \State $\mathrm{SoC}_{\text{new}} \gets \mathrm{SoC} - kWh_{\text{bat}} \cdot \eta$
        \State $\text{buy\_kWh} \gets \text{deficit} - kWh_{\text{bat}}$
    \EndIf
\ElsIf{$a_t = 5$}
    \State \textit{No state change.}
\ElsIf{$a_t = 6$}
    \If{$G > L$ and $\mathrm{SoC} \leq 80$ and $\tau \neq \text{P}$}
        \State $\text{surplus} \gets G-L$
        \State $kWh_{\text{bat}} \gets \min(C_{\text{CHARGE}}, \text{surplus})$
        \State $\mathrm{SoC}_{\text{new}} \gets \min(\mathrm{SoC} + kWh_{\text{bat}} \cdot \eta, 100)$
    \EndIf
\ElsIf{$a_t = 7$}
    \If{$G < L$ and $\mathrm{SoC} \geq 20$}
        \State $\text{need} \gets L-G$
        \State $kWh_{\text{bat}} \gets \min(C_{\text{CHARGE}}, \text{need}, \mathrm{SoC} \cdot C_{\text{BAT}} / 100)$
        \State $\mathrm{SoC}_{\text{new}} \gets \mathrm{SoC} - kWh_{\text{bat}} \cdot \eta$
        \State $\text{buy\_kWh} \gets \max(\text{need} - kWh_{\text{bat}}, 0)$
    \EndIf
\EndIf
\State $\mathrm{SoC}_{\text{new}} \gets \max(0, \min(\mathrm{SoC}_{\text{new}}, 100))$
\State \textbf{Output:} $s_{t+1} = [L_{t+1}, G_{t+1}, \mathrm{SoC}_{\text{new}}, (h+1) \bmod 24, \mathrm{ISP}_{t+1}, \mathrm{IBP}_{t+1}]$
\end{algorithmic}

\section{}
\label{hyperparameters}

\begin{figure*}
    \centering
    \includegraphics[width=1\textwidth]{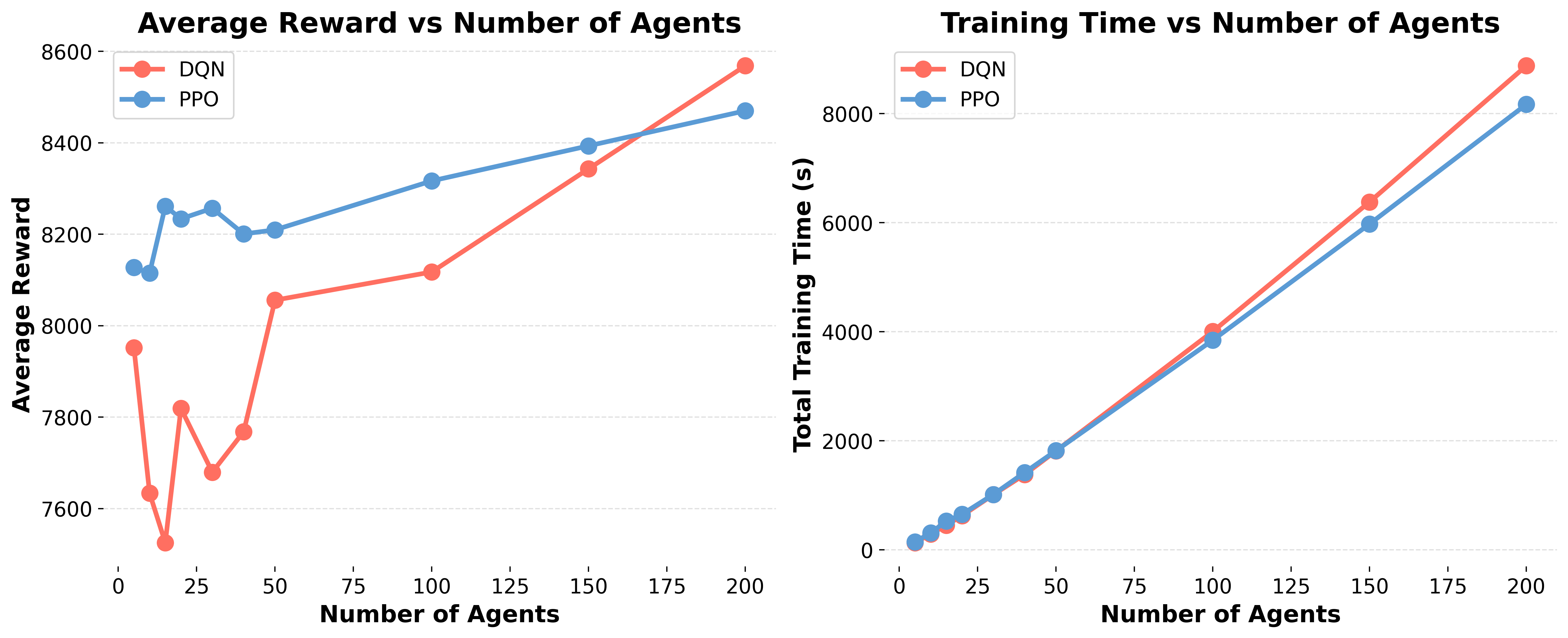}
    \caption{Average reward and total training time for PPO and DQN across varying numbers of agents}
    \label{scalabilityandsensitivityfigure}
\end{figure*}

This appendix provides the default hyperparameters for the PPO and DQN algorithms, outlined in Tables~\ref{ppo_params} and~\ref{dqn_params}, as implemented in the Stable-Baselines3 library and tailored for multi-agent environments, as well as the scalability and sensitivity analysis of hyperparameters. The table for sensitivity analysis is Table~\ref{sensitivitytable}, for scalability is Table~\ref{scalabilitytable}, and the plots for average reward and total training time for PPO and DQN across varying numbers of agents are shown in Figure~\ref{scalabilityandsensitivityfigure}.

\begin{table}
\centering
\begin{tabular}{llp{8cm}}
\hline
\textbf{Parameter} & \textbf{Default Value} & \textbf{Description} \\
\hline
Learning Rate & 0.0003 & Step size for optimizer updates \\

n\_steps & 2048 & Number of steps per environment per update \\

Batch Size & 64 & Minibatch size for optimization \\

gamma & 0.99 & Discount factor for future rewards \\

gae\_lambda & 0.95 & GAE (Generalized Advantage Estimation) parameter \\

clip\_range & 0.2 & Clipping parameter for policy loss \\

clip\_range\_vf & None & Value function clipping parameter \\

normalize\_advantage & True & Flag for advantage normalization \\

ent\_coef & 0.0 & Entropy coefficient for exploration \\

vf\_coef & 0.5 & Value function coefficient \\

max\_grad\_norm & 0.5 & Maximum norm for gradient clipping \\

use\_sde & False & State-dependent exploration flag \\

sde\_sample\_freq & -1 & SDE sampling frequency \\

Policy & MlpPolicy & Default neural network policy type \\
\hline
\end{tabular}
\caption{Default PPO Hyperparameters in Stable-Baselines3}
\label{ppo_params}
\end{table}

\begin{table}
\centering
\begin{tabular}{llp{8cm}}
\hline
\textbf{Parameter} & \textbf{Default Value} & \textbf{Description} \\
\hline
Learning Rate & 0.0001 & Step size for optimizer updates \\

Buffer Size & 1,000,000 & Replay buffer capacity \\

Learning Starts & 50000 & Steps before learning begins \\

Batch Size & 32 & Minibatch size for training \\

Tau & 1.0 & Soft update coefficient \\

gamma & 0.99 & Discount factor \\

train\_freq & 4 & Model update frequency \\

Gradient Steps & 1 & Number of gradient steps per update \\

Target Update Interval & 10000 & Target network update frequency \\

exploration\_fraction & 0.1 & Exploration rate decay period \\

exploration\_initial\_eps & 1.0 & Initial exploration rate \\

exploration\_final\_eps & 0.05 & Final exploration rate \\

max\_grad\_norm & 10 & Maximum norm for gradient clipping \\

Policy & MlpPolicy & Default neural network policy type \\
\hline
\end{tabular}
\caption{Default DQN Hyperparameters in Stable-Baselines3}
\label{dqn_params}
\end{table}

\begin{table}
\centering
\caption{Parameter Sensitivity Results for PPO and DQN}
\begin{tabular}{lcccc}
\hline
\textbf{Algorithm} & \textbf{Parameter} & \textbf{Value} & \textbf{Avg Reward} & \textbf{Std Reward} \\
\hline
\multirow{3}{*}{PPO} & Learning Rate & 0.0003 & 8148.3 & 184.5 \\
 & Learning Rate & 0.001 & 8183.7 & 128.5 \\
 & Learning Rate & 0.003 & 8058.4 & 234.2 \\

 & Batch Size & 32 & 8204.7 & 161.2 \\
 & Batch Size & 64 & 7411.2 & 2190.0 \\
 & Batch Size & 128 & 8067.2 & 303.9 \\
\hline
\multirow{3}{*}{DQN} & Learning Rate & 0.0003 & 4515.1 & 1816.9 \\
 & Learning Rate & 0.001 & 6939.9 & 1248.7 \\
 & Learning Rate & 0.003 & 7214.1 & 594.0 \\

 & Batch Size & 32 & 4401.2 & 1239.8 \\
 & Batch Size & 64 & 5237.2 & 1529.9 \\
 & Batch Size & 128 & 7048.0 & 694.6 \\
\hline
\end{tabular}
\label{sensitivitytable}
\end{table}

\begin{table}
\centering
\caption{Scalability Results for PPO and DQN}
\begin{tabular}{lccccc}
\hline
\textbf{Alg.} & \textbf{Agents} & \textbf{Avg Reward} & \textbf{Std Reward} & \textbf{Training Time (s)} & \textbf{Bandwidth (kbps)} \\
\hline
\multirow{6}{*}{PPO} & 5 & 8127.2 & 181.7 & 149.1 & 12.5 \\
 & 10 & 8115.0 & 204.5 & 311.2 & 25 \\
 & 20 & 8233.8 & 313.5 & 653.3 & 50 \\
 & 50 & 8209.6 & 351.2 & 1820.2 & 125 \\
 & 100 & 8316.7 & 528.5 & 3846.3 & 250 \\
 & 200 & 8470.1 & 861.3 & 8168.6 & 500 \\
\hline
\multirow{6}{*}{DQN} & 5 & 7951.6 & 370.1 & 132.6 & 12.5 \\
 & 10 & 7633.4 & 1654.1 & 289.5 & 25 \\
 & 20 & 7819.1 & 1817.5 & 628.0 & 50 \\
 & 50 & 8055.8 & 1200.2 & 1816.1 & 125 \\
 & 100 & 8117.5 & 2240.0 & 4005.8 & 250 \\
 & 200 & 8568.5 & 3370.2 & 8876.5 & 500 \\
\hline
\end{tabular}
\label{scalabilitytable}
\end{table}

\end{document}